\documentclass[10pt,twocolumn,letterpaper]{article}

\usepackage[pagenumbers]{cvpr} % To force page numbers, e.g. for an arXiv version

\usepackage[accsupp]{axessibility} 
\usepackage{graphicx}
\usepackage{amsmath}
\usepackage{amssymb}
\usepackage{booktabs}

\usepackage{pifont} % for cmark and xmark
\newcommand{\cmark}{\ding{51}}%
\newcommand{\xmark}{\ding{55}}%

\usepackage{graphicx}
\usepackage{booktabs}
\usepackage{multirow}
\usepackage{makecell}
\usepackage{caption}
\usepackage{threeparttable}
\usepackage{algorithm} 
\usepackage{algorithmic} 
\usepackage{listings}
\usepackage{bm}
\usepackage[dvipsnames]{colortbl}
\usepackage[dvipsnames]{xcolor} 
\definecolor{Gray}{gray}{0.85}
\usepackage{hhline}
\def\eg{\textit{e.g.}}
\def\ie{\textit{i.e.}}

\def\etal{\textit{et al.}}

\definecolor{KleinBlue}{rgb}{0.0, 0.129, 0.7}

\newcommand{\methodname}{AdaTAD}

\definecolor{cvprblue}{rgb}{0.21,0.49,0.74}
\usepackage[pagebackref,breaklinks,colorlinks,citecolor=cvprblue]{hyperref}

\usepackage[capitalize]{cleveref}
\crefname{section}{Sec.}{Secs.}
\Crefname{section}{Section}{Sections}
\Crefname{table}{Table}{Tables}
\crefname{table}{Tab.}{Tabs.}

\def\paperID{****} % *** Enter the Paper ID here
\def\confName{CVPR}
\def\confYear{2024}

\title{End-to-End Temporal Action Detection with 1B Parameters Across 1000 Frames  
}

\author{
Shuming Liu$^{1}$ \quad
Chen-Lin Zhang$^{2}$ \quad
Chen Zhao$^{1}$\thanks{Corresponding author.} \quad
Bernard Ghanem$^{1}$
\and
$^{1}$King Abdullah University of Science and Technology (KAUST)
\quad $^{2}$4Paradigm Inc
}

\begin{document}
\maketitle
\begin{abstract}

Recently, temporal action detection (TAD) has seen significant performance improvement with end-to-end training. However, due to the memory bottleneck, only models with limited scales and limited data volumes can afford end-to-end training, which inevitably restricts TAD performance. 
In this paper, we reduce the memory consumption for end-to-end training, and manage to scale up the TAD backbone to \textbf{1 billion parameters} and the input video to \textbf{1,536 frames}, leading to significant detection performance.
The key to our approach lies in our proposed temporal-informative adapter (TIA), which is a novel lightweight module that reduces training memory. Using TIA, we free the humongous backbone from learning to adapt to the TAD task by only updating the parameters in TIA. TIA also leads to better TAD representation by temporally aggregating context from adjacent frames throughout the backbone.
We evaluate our model across four representative datasets. Owing to our efficient design, we are able to train end-to-end on VideoMAEv2-giant and achieve 75.4\% mAP on THUMOS14, being the first end-to-end model to outperform the best feature-based methods. Code is available at \url{https://github.com/sming256/AdaTAD}.

\end{abstract}    
\section{Introduction}
\label{sec:intro}

Temporal Action Detection (TAD) plays a vital role in the understanding of long-form videos. Its objective is to pinpoint specific action instances within untrimmed videos, identifying their start and end times, along with their respective categories~\cite{caba2015activitynet,escorcia2016DAPsDA,zhao2019hacs,yang2023basictad, xu2020g,ramazanova2023owl,zhao2021video}. This task is crucial for various applications, including highlight detection~\cite{yao2016highlight,mai2023egoloc}, video-language grounding~\cite{Mun_2020_CVPR,Soldan_2021_ICCV,soldan2021mad}, and action spotting~\cite{Alwassel2017ActionSS}.
Though innovations in the detector design have made profound progress in the past years~\cite{lin2019bmn,zhao2021video, zhang2022actionformer}, recent research highlights two new trends: \textit{end-to-end training}~\cite{cheng2022tallformer,zhao2023re2tal,liu2023etad}, and \textit{scaling up}~\cite{wang2022internvideo,wang2023videomae}.

\begin{figure}[t]
\centering
\includegraphics[width=1.0\linewidth]{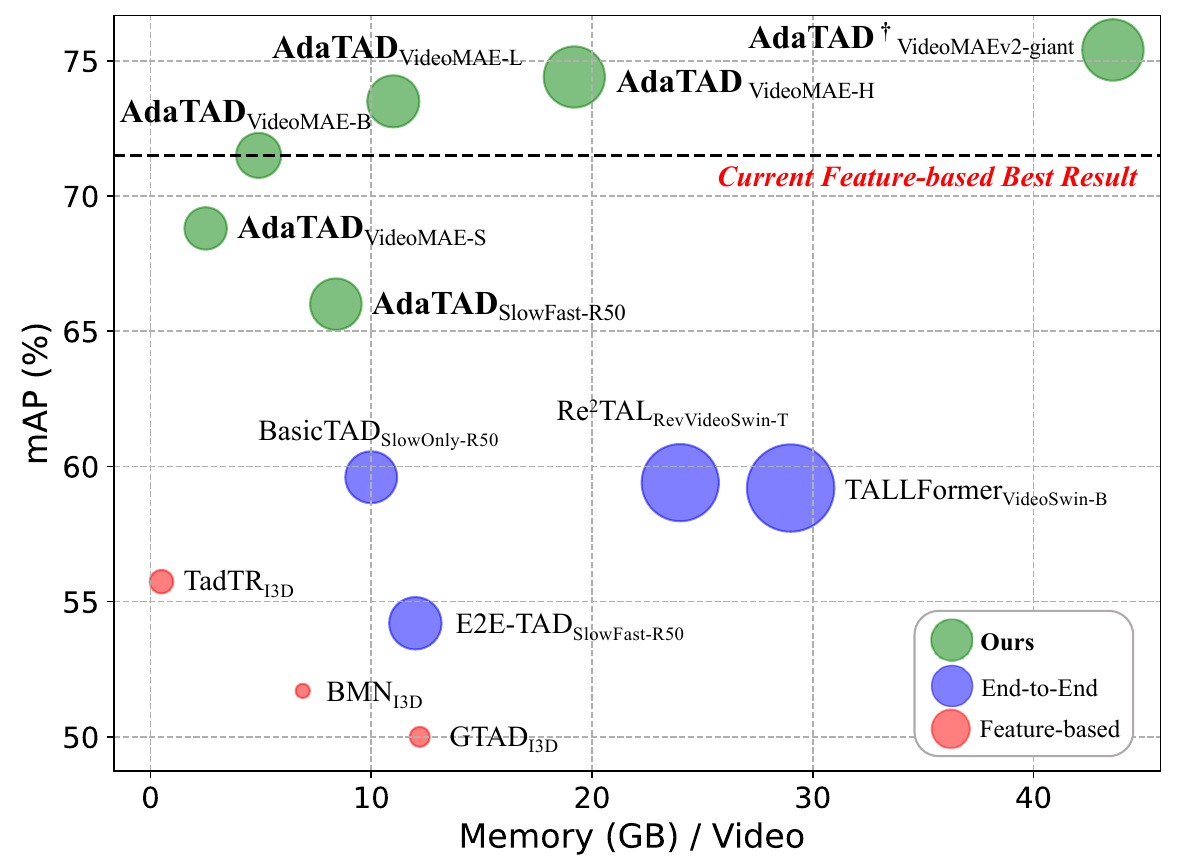}
\vspace{-18pt}
\caption{\textbf{\methodname{} enjoys the benefit of end-to-end training and scaling up with efficient memory usage.}
The bubble size represents the number of the model's learnable parameters. Using SlowFast-R50, AdaTAD achieves better performance compared to E2E-TAD~\cite{liu2022empirical} with less memory. When further scaling up the model to VideoMAEv2-gaint and data to 1536 frames, we achieve an impressive avg. mAP of 75.4\% on THUMOS14.}
\label{fig:intro}
\vspace{-12pt}
\end{figure}

\textit{End-to-end training} in TAD refers to jointly training the video encoder and action detector~\cite{lin2021learning,yang2023basictad,cheng2022tallformer,zhao2024dr2net}.
Compared to feature-based methods, end-to-end training offers distinct advantages. First, it can effectively bridge the gap commonly found between pretraining and fine-tuning, such as data and task discrepancy. Second, video spatial augmentations can be utilized in the end-to-end setting, leading to further performance gain.

\textit{Scaling up} refers to improving performance by increasing the model size or the input data volume, and has demonstrated its effectiveness in various domains~\cite{dosovitskiy2020image,he2022masked,tong2022videomae,openai2023gpt4}. In TAD, offline methods have attempted to scale up the feature extraction network to reach a higher performance. A notable example includes the work by Wang \etal \cite{wang2023videomae}, which reports a 10\% increase in mean Average Precision (mAP) by scaling from VideoMAE-S to VideoMAEv2-giant, using ActionFormer~\cite{zhang2022actionformer} on THUMOS14~\cite{jiang2014thumos}.

Intuitively, combining the strengths of end-to-end training and scaling up is expected to be most beneficial for improving TAD performance. However, both strategies demand substantial GPU memory, which restricts end-to-end training to a small model~\cite{zhao2023re2tal,lin2021learning,yang2023basictad}, or a small input volume~\cite{liu2022empirical,cheng2022tallformer}. As shown in Fig.~\ref{fig:intro}, the performance of previous end-to-end methods still significantly lags behind the best results achieved by feature-based approaches. Additionally, current end-to-end methods in TAD use computationally intensive full fine-tuning, risking the issues of catastrophic forgetting and overfitting during transfer learning~\cite{sung2022vl,yang2023aim}. These issues can result in less competitive performance, especially when the downstream datasets are small, which is a common scenario in the TAD domain. 

In this paper, we aim to overcome the above limitations by harnessing the advantages of both scaling-up and end-to-end training. To achieve this, we introduce adapter tuning for temporal action detection  \textbf{(\methodname{})}. Our method successfully trains a TAD model in an end-to-end manner, utilizing a backbone with 1 billion parameters and processing input videos of 1,536 frames. As illustrated in Fig.~\ref{fig:intro}, to the best of our knowledge, this is the first end-to-end work that outperforms the best feature-based TAD methods.

Specifically, we employ the following strategies to enhance the TAD performance while maintaining reasonable memory consumption.
\textbf{First}, we identify that the snippet representation commonly used in feature-based methods is excessively redundant. In response, we have adopted a more memory-efficient frame-representation scheme, establishing an effective end-to-end baseline for TAD. \textbf{Second}, 
we adopt the parameter-efficient fine-tuning technique to minimize memory usage and mitigate overfitting in transfer learning. Notably, we introduce a novel Temporal-Informative Adapter (TIA). This adapter is injected between backbone layers and is the only learnable component in the backbone during fine-tuning. Different from conventional adapters~\cite{houlsby2019parameter}, TIA is tailored for the TAD task and integrates temporal depth-wise convolutions to aggregate informative context from adjacent frames.
\textbf{Third}, for additional memory efficiency, we propose a lighter variant of our method. By positioning the TIAs alongside the original backbone, rather than inside it, backpropagation can be done through the TIAs only. This configuration can further cut down on the need for intermediate activations within the primary backbone, thereby allowing us to scale up the model size and data size to unprecedented levels.

\methodname{} establishes a new state-of-the-art across multiple TAD datasets.
Notably, our method achieves an impressive 75.4\% mAP on THUMOS14, outperforming the previous feature-based best result of 71.5\% by a large margin. This achievement underscores the possible paradigm shift in TAD, \ie, moving from traditional feature extraction plus offline detectors to scaling up end-to-end TAD training. We summarize our contribution as follows:

\begin{enumerate}
\item We introduce an efficient end-to-end framework for TAD, scaling up the model size to 1 billion parameters and the input data to 1,536 frames. We achieve a consistent performance improvement with the scaling up, shedding light on the importance of scaling for TAD.

\item  We propose a novel temporal-informative adapter to reduce memory as well as aggregate the temporal context for TAD. Different variants of these adapters are designed to trade off between performance and memory cost. To the best of our knowledge, we are the first to introduce the adapter mechanism to TAD.
\item  Our method achieves state-of-the-art performance across four TAD datasets. Remarkably, this represents the first end-to-end approach that outperforms the previous feature-based methods by a large margin.
\end{enumerate}

%-------------------------------------------------------------------------

\section{Related Work}

\begin{figure*}[t]
\centering
\includegraphics[width=0.96\linewidth]{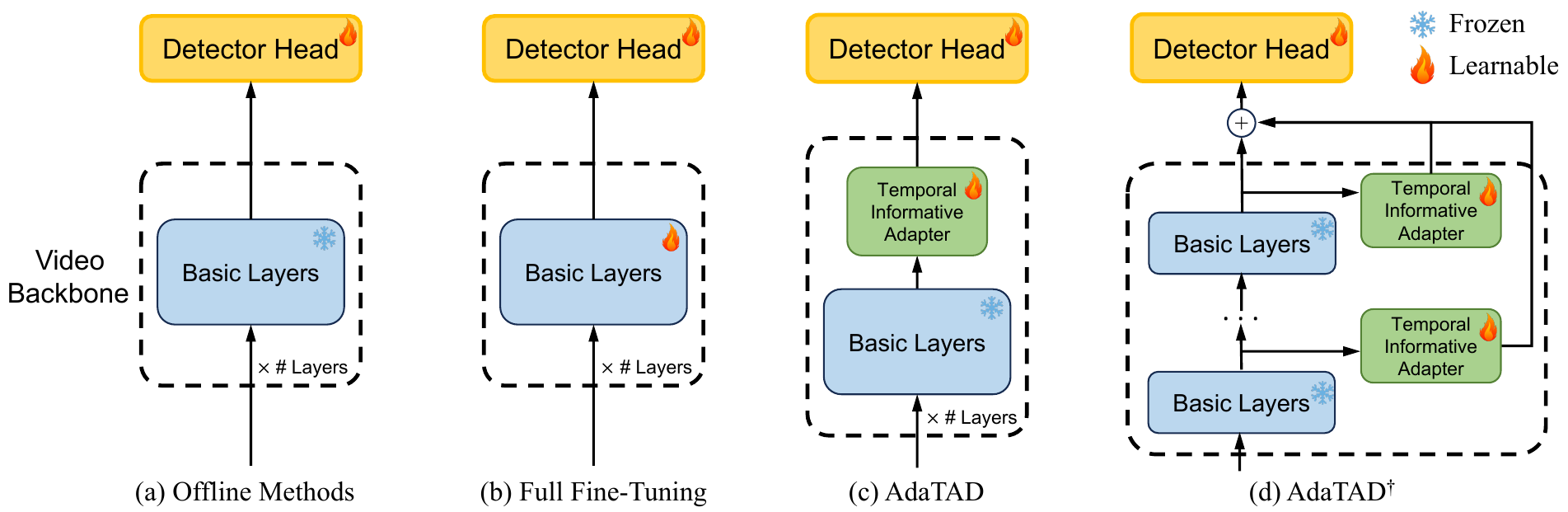}
\vspace{-4pt}
\caption{\textbf{Comparative illustration of our proposed TAD framework versus popular and widely used alternatives.} (a) represents the typical offline method. (b) is the traditional end-to-end method using full fine-tuning. (c) Tailored for the TAD task, our \methodname{} uses a lightweight temporal-informative adapter inside the backbone to achieve efficient transfer learning. (d) To further reduce memory usage and scale up the model/data, \methodname{}$^\dagger$ uses an alternative placement for adapters outside the backbone.}
\label{fig:framework}
\vspace{-8pt}
\end{figure*}

\noindent{\textbf{Temporal Action Detection.}} Current methods for temporal action detection, also referred to as temporal action localization, can be broadly classified into three categories based on their architectural design: one-stage, two-stage, and DETR-based methods. One-stage methods directly localize actions from a multi-scale feature pyramid, such as ActionFormer~\cite{zhang2022actionformer} and TriDet~\cite{shi2023tridet}. These methods integrate action classification and temporal boundary regression in a single step~\cite{yang2020revisiting,yang2023basictad,shao2023action}. Two-stage methods, in contrast, involve an additional step of proposal feature extraction \cite{lin2019bmn,qing2021temporal,Bai2020bcgnn,zhao2022segtad,wang2020boundary,xia2022learning,Zhao_2023_ICCV}. For instance, VSGN~\cite{zhao2021video} employs boundary sampling to further refine proposals. Recently, there is a growing interest in query-based methods~\cite {tan2021relaxed,liu2022end,shi2022react}, which deploy a set of learned queries to interact with the feature maps and directly predict the actions' temporal boundaries and categories.

In addition to the aforementioned categories, TAD can also be divided into feature-based and end-to-end methods. The former relies on pre-extracted RGB features and optionally incorporates optical flow features. On the other hand, end-to-end methods take raw video frames as input and jointly optimize the video encoder and action detector \cite{liu2020progressive}. Due to computational constraints, AFSD~\cite{lin2021learning} downsamples the input to a resolution of $96^2$. DaoTAD~\cite{wang2021rgb} and E2E-TAD~\cite{liu2022empirical} provide evidence that high TAD performance can be achieved by relying solely on the RGB modality with various data augmentations. Further innovations came from SGP~\cite{cheng2022stochastic}, TALLFormer~\cite{cheng2022tallformer}, and ETAD~\cite{liu2023etad}, all of which introduced strategies to backpropagate only through parts of the data. Additionally, Re$^2$TAL~\cite{zhao2023re2tal} and Dr$^2$Net\cite{zhao2024dr2net} design reversible network architectures to release the memory occupied by intermediate activations. Despite these advancements, all above methods follow the full fine-tuning paradigm, and none has yet surpassed the best results achieved by feature-based approaches. 

\vspace{2pt}
\noindent{\textbf{Scaling Law in Deep Learning.}} Scaling up model and data has been a prevalent strategy across both computer vision and natural language processing fields to achieve superior performance. The GPT series~\cite{radford2018improving, radford2019language,brown2020language,openai2023gpt4} has consistently demonstrated that larger models pretrained on extensive datasets yield significant gains in language understanding capabilities. Analogously, in the realms of image and video understanding, architectures such as ViT~\cite{dosovitskiy2020image} and MViT~\cite{fan2021multiscale}, have also witnessed the effectiveness of this scaling strategy. Alabdulmohsin \etal \cite{alabdulmohsin2022revisiting} present a recipe for estimating scaling law parameters reliably from learning curves in computer vision. 
To attain even greater performance, several studies have also scaled up image resolution~\cite{liu2021swin} or video clip length~\cite{wang2023videomae}. In this paper, we successfully apply this principle within the domain of temporal action detection and achieve state-of-the-art results.

\vspace{2pt}
\noindent{\textbf{Efficient Transfer Learning.}} Transfer learning aims to adapt a pretrained model to a new domain. In TAD, typically off-the-shelf action recognition models are employed as the backbone, such as SlowFast~\cite{slowfast_iccv19}. Traditional transfer learning adopts full fine-tuning, meaning that all parameters of the pretrained model are updated. However, studies by \cite{radford2021learning,yang2023aim} have noted that full fine-tuning may harm the pre-learned knowledge, particularly when the downstream dataset is small and less comprehensive. Moreover, as models increase in size, the computational and storage demands of full fine-tuning proportionally increase.

In response to these challenges, several works have investigated parameter-efficient tuning (PEFT) strategies that involve fine-tuning only a fraction of the network.
For instance, 
Adapter~\cite{houlsby2019parameter} inserts lightweight modules analogous to feedforward networks in transformers, and only tunes these elements. LoRA~\cite{hu2021lora} employs low-rank matrices in each transformer layer. Prefix-tuning~\cite{li2021prefix} and prompt-tuning~\cite{lester2021power} attach learnable prompt tokens at the input stage or within each layer. In computer vision, many PEFT methods \cite{jia2022visual,chen2022conv, chen2022adaptformer,yang2023aim} have also been explored across various tasks to optimize transfer learning efficiency. Our work represents the first effort to examine the potential of the PEFT mechanism in TAD.

Although PEFT effectively reduces the number of learnable parameters, data-intensive and computationally heavy tasks like video understanding require more memory-efficient techniques. 
To this end, several works try to externalize the trainable components from the backbone, eliminating the need for backpropagation through the extensive original model.
For example, LST~\cite{sung2022lst} introduces a supplementary lightweight network that operates in parallel with the main model. Similarly, E$^3$VA~\cite{yin2023parameter} leverages intermediate features with adapters to enable efficient transfer learning while minimizing memory usage. Our work is inspired by these methods yet with a streamlined and simple design.

\section{Methodology}

In this section, we introduce our AdaTAD step-by-step. We first introduce notations and study the efficient video representation to establish an end-to-end TAD baseline. Next, we introduce a temporal-informative adapter designed for efficient TAD. Finally, we propose an alternative placement for adapters to further alleviate computational demands.

\subsection{Notations}

Temporal action detection can be formulated as follows: given an untrimmed video $\mathbf{X} \in \mathbb{R}^{3\times H \times W \times T}$, where $H$ and $W$ are the height and width of each frame, and $T$ is the frame number, its temporal action annotations can be denoted as ${{\Psi }_{g}}=\left\{\varphi _i\!=\!(t_s,t_e,c)\right\}_{i=1}^N$, where ${t_s},{t_e},c$ are the start, end time and category of action instance ${{\varphi }_{i}}$, and $N$ is the number of ground truth actions. TAD aims to predict candidate proposal set ${{{\Psi }_{p}}=\left\{\hat{\varphi}_i\!=\!({\hat{t}_s},\hat{t}_e,\hat{c},s) \right\}_{i=1}^{{M}}}$ to cover ${{\Psi }_{g}}$, and $s$ is the confidence score.

\subsection{Frame-level representation}
\label{sec:e2e_baseline}
Our end-to-end TAD architecture comprises two main components: feature extraction and action detection. Following previous work~\cite{zhao2023re2tal}, we select ActionFormer~\cite{zhang2022actionformer} as our action detection head due to its robust performance across various datasets without much hyperparameter tuning. 
Next, we discuss two ways of encoding raw frames into representative features (feature extraction): snippet representation and frame representation.

\textbf{Snippet Representation.} Snippet-based video representations are popular choices in offline feature extraction. The whole video is divided into several short snippets (or namely clips). Each snippet has a short temporal length, \eg, 16 frames, and different snippets can have overlapping frames. Thus, the video can be conceptualized as $T$ snippets, denoted by $\mathbf{X} \in \mathbb{R}^{T \times 3 \times 16 \times H \times W}$. Each snippet is processed through the video backbone, and spatial-temporal pooling is applied to extract one snippet feature. 
This processing yields the feature representation $\mathbf{F} \in \mathbb{R}^{T \times C}$, where $C$ denotes the channel dimension of the pooled features. 

\textbf{Frame Representation.} In contrast to snippet-based representations, frame-based video representations consider the entire video as a singular snippet or clip, represented as $\mathbf{X} \in \mathbb{R}^{1 \times 3  \times T\times H \times W}$. Then, the whole frame sequence is fed into the video backbone, and only spatial pooling is employed to extract features \cite{lin2021learning,cheng2022tallformer,zhao2023re2tal}.
For attention-based models such as VideoMAE~\cite{vaswani2017attention}, the video is chunked into multiple shorter clips to avoid extensive temporal attention. 

\begin{table}[b]
\caption{\textbf{Snippet representation \textit{vs} frame representation.} We use the end-to-end version of ActionFormer with two representations for comparison. The snippet input is $768\! \times \!3  \!\times \!16 \!\times\! 160\! \times \!160$, and the frame input is $1\! \times \!3  \!\times \!768 \!\times\! 160\! \times \!160 $. $*$ means activation checkpointing is utilized to avoid overflowing GPU memory.}
\vspace{-2pt}
\centering
\small
\setlength{\tabcolsep}{5pt}
\begin{tabular}{l|c|ccc}
    \toprule
        \textbf{Setting}   & \textbf{Backbone}  & \textbf{Repr.} & \textbf{Avg. mAP}  & \textbf{Mem (GB)} \\
    \midrule
    \multirow{4}{*}{Frozen }  
    & \multirow{2}{*}{VideoMAE-S}  & Frame & 59.35 & 1.9    \\
    & & Snippet & 57.68 & 13.2 \\
    \cmidrule{2-5}
    & \multirow{2}{*}{SlowFast-R101}  & Frame & 61.34 & 3.6   \\
    & & Snippet & 60.24 & 17.2 \\
    \midrule
    \multirow{4}{*}{\makecell{End \\to \\End }}  
    & \multirow{2}{*}{VideoMAE-S}  & Frame & 67.15 & 2.8$^*$    \\
    & & Snippet & 68.46 & 24.6$^*$ \\
    \cmidrule{2-5}
    & \multirow{2}{*}{SlowFast-R101}  & Frame & 65.33 & 5.5$^*$   \\
    & & Snippet & 66.72 & 51.6$^*$ \\
    \bottomrule
\end{tabular}
\label{tab:ablation_representation}
% \vspace{-10pt}
\end{table}

Although both representations have been employed in previous studies, a fair comparison between them has not yet been performed. To address this gap, we conduct a comparative analysis of the two representations under the same setting on THUMOS14, measuring their memory usage and detection mAP. 
The results in Table~\ref{tab:ablation_representation} indicate that \textbf{frame representation has comparable or even better performance than snippet representation}, yet with much smaller memory consumption. When the feature extraction backbones are frozen, frame representation yields superior results to snippet representation for both VideoMAE~\cite{tong2022videomae} and SlowFast~\cite{slowfast_iccv19} backbones. Only in the end-to-end setting can the snippet representation achieve 1\% mAP advantage over frame representations; however, it requires 8 times more memory consumption. Taking into account both performance and memory usage, frame-based representations could be a better choice for end-to-end TAD development. 

Therefore, we use frame representation as the default baseline to encode videos in our experiments. Following the previous TALLFormer work~\cite{cheng2022tallformer}, we also incorporate activation checkpointing~\cite{chen2016training} and mixed precision training \cite{micikevicius2017mixed} to fully harness the potential of scaling.

\subsection{Temporal-Informative Adapter}
\label{sec:ti_adapter}

In Section~\ref{sec:e2e_baseline}, we have built a simple end-to-end baseline using full fine-tuning. However, the baseline still suffers from two aspects: \textbf{1. Increased computational cost.} In Table~\ref{tab:ablation_representation}, we only use small video backbones like VideoMAE-S. When scaling VideoMAE-S to larger models, the computational burden and memory cost will grow rapidly. \textbf{2. Inferior transfer learning ability.} More critically, the baseline follows the tradition of full fine-tuning, which may lead to inferior transfer learning. Pointed out by~\cite{sung2022vl,yang2023aim}, full fine-tuning may result in overfitting or forgetting, especially for large pretrained models. If downstream datasets are not sufficiently diverse, full fine-tuning can even destroy the powerful features learned from large-scale pretraining. 
Motivated by the above two aspects, we apply the PEFT mechanism and propose to fine-tune a plug-and-play module named \textbf{Temporal-Informative Adapter (TIA)} to achieve efficient and effective transfer learning for TAD. 

We first review the architecture of the standard adapter proposed by ~\cite{houlsby2019parameter}.
As formulated in Equation~\ref{eq:adapter_ori}, the standard adapter includes a down-projection fully connected (FC) layer with parameter  $\bm W_{\rm down} \in \mathbb{R}^{d \times \frac{d}{\gamma}}$, where $\frac{d}{\gamma}$ represents the intermediate dimension and satisfies $\gamma \!>\! 1 $.  Then, an up-projection layer $\bm W_{\rm up} \in \mathbb{R}^{\frac{d}{\gamma} \times d}$ is employed to restore the channel dimension. Between these two FC layers, a non-linear activation function $\bm {\sigma}$ is inserted, such as GELU~\cite{hendrycks2016gaussian}. Afterward, a residual connection is added to the output of the projection layer. Note that $\bm{x}$ and $\bm{x'}$ are the input and output features with the same shape $\mathbb{R}^{d\times t\times h\times w}$.
\begin{equation}    
    \bm{x'} = {\bm W_{\rm up}^{\rm \top}} \cdot \bm {\sigma} ({\bm W_{\rm down}^{\rm \top}} \cdot \bm{x} ) + \bm{x}.
    \label{eq:adapter_ori}
\end{equation}

\begin{figure}[t]
\centering
\includegraphics[width=0.98\linewidth]{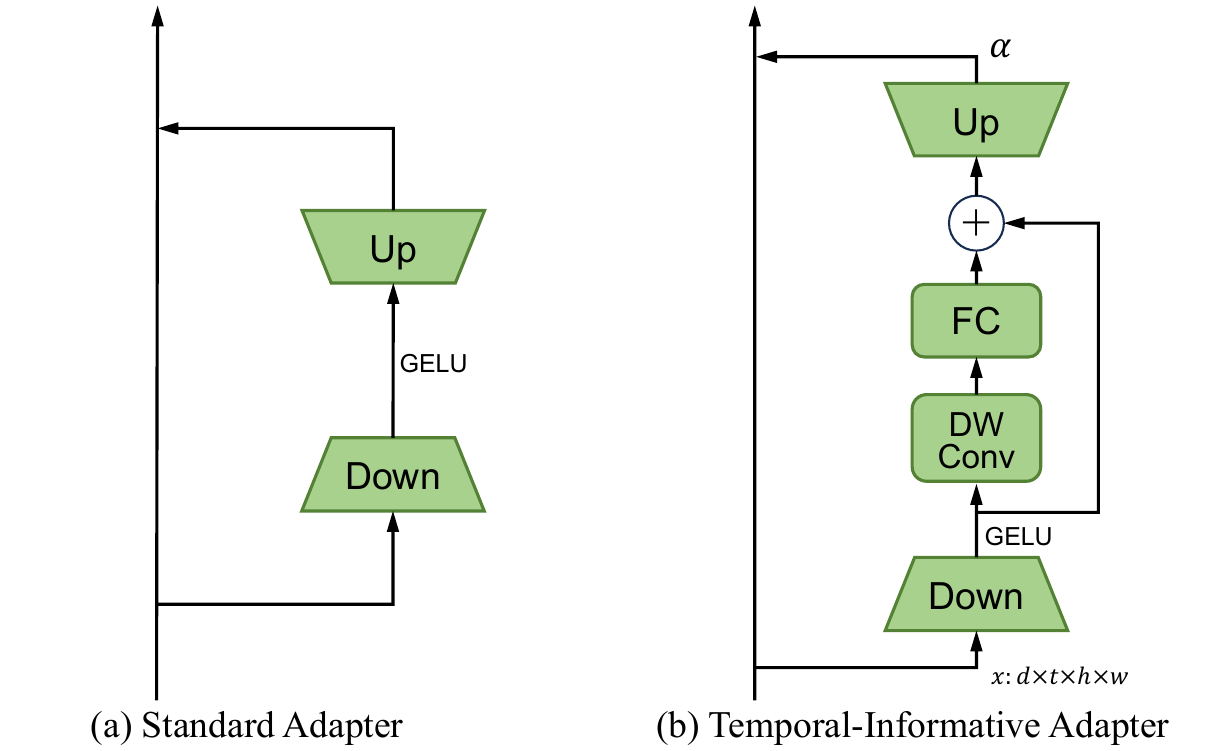}
\vspace{-6pt}
\caption{\textbf{Architecture of (a) standard adapter and (b) our temporal-informative adapter}. We incorporate temporal depth-wise convolution to aggregate context from adjacent frames.}
\label{fig:adapter_arch}
% \vspace{-12pt}
\end{figure}

Although the adapter has achieved great success in natural language processing and computer vision, the standard adapter only focuses on adapting channel information, which neglects the temporal context vital for the TAD task. To address this limitation, we introduce the temporal-informative adapter, as depicted in Fig.~\ref{fig:adapter_arch}(b).

The architecture of TIA follows the general bottleneck design of the standard adapter, while we integrate the temporal depth-wise convolution layers, as described in Equation~\ref{eq:our_adapter}. The temporal convolution with a kernel size of $k$ is designed to aggregate local informative context from adjacent frames and to enrich the representation of the current time step. Practically, this is achieved through the application of a 3D convolution with kernel size $(k,1,1)$ and group size $\frac{d}{\gamma}$ for depth-wise processing.
Additionally, an FC layer with weight $\bm W_{\rm mid} \in \mathbb{R}^{\frac{d}{\gamma} \times \frac{d}{\gamma}}$ is employed to facilitate information exchange across channels. At last, a learnable scalar $ \alpha$ is introduced to adjust the amplitude of adapter's output.
\begin{equation}
\begin{aligned}
 \bm{\bar x} &= \bm {\sigma} ({\bm W_{\rm down}^{\rm \top}} \cdot \bm{x} ), \\
 \bm{\hat x} &= {\bm W_{\rm mid}^{\rm \top}} \cdot \operatorname{\textbf{DWConv}}_k(\bm{\bar x}) + \bm{\bar x}, \\
 \bm{x'}  &= \alpha \cdot {\bm W_{\rm up}^{\rm \top}} \cdot \bm{\hat x}  + \bm {x} .
\end{aligned}
\label{eq:our_adapter}
\end{equation}

As shown in Fig.~\ref{fig:framework}(c), TIA is designed to be inserted between different backbone layers, \eg~ between each ViT block of VideoMAE or each bottleneck block of SlowFast. To ensure the newly added connection does not affect the original network at the beginning of transfer learning, the weight and bias of the adapter's last projection layer $\bm W_{\rm up}$ are initialized to 0. The learnable coefficient ${\alpha}$ is initialized to 1. The temporal kernel size $k$ is set to 3, and the channel downsampling ratio $\gamma$ is set to 4 by default. Under these settings, the additional trainable parameters coming from TIA only account for 4.7\% of the total parameters of the original backbone. Since this backbone is frozen when TIA is used, our proposed strategy constitutes a massive reduction in trainable parameters as compared to full fine-tuning. Our experiments show that TIA can achieve better performance than full fine-tuning with less memory usage.

\subsection{Alternative Placement for Adapter}

Although the previously described PEFT approach can reduce tunable parameters and memory usage, the gradient still needs to backpropagate over the entire backbone during training. This requirement limits our ability to scale-up the model size or input data size further. As highlighted in prior works~\cite{sung2022lst,yin2023parameter}, if we can stop the gradient backpropagation within the original backbone, additional computational savings can be achieved.

Inspired by this insight, we propose a placement strategy for adapters that position them externally to the backbone, rather than inserting them inside. As illustrated in Fig.~\ref{fig:framework}(d), we utilize the previously introduced TIA module, but its output does not feed back into the middle of the original backbone. It is directly added to the backbone's final layer. This configuration eliminates the need for backpropagation through the original network, as gradients are only tracked to the shallow lightweight adapter. To further diminish computation, we observe that adapting only the last half of backbone layers yields comparable performance while reducing half of the adaptation cost.

To distinguish the different variants, we name the standard adaption design as \methodname{}, and the alternative placement as \methodname{}$^{\dagger}$. The latter can be considered as a lite version of the former. 
Compared to directly injecting adapters into the backbone, \methodname{}$^{\dagger}$ may lead to a slight performance drop. However, it enables us to leverage richer models and more data, which should effectively counter this possible drop. 

\section{Experiments}

\begin{table*}[t]
\centering
\caption{\textbf{Results on ActivityNet-1.3 and THUMOS14}, measured by mAP (\%) at different tIoU thresholds. E2E refers to end-to-end training, and Mem refers to memory usage (GB) per video. On ActivityNet-1.3, our prediction is combined with CUHK~\cite{zhao2017cuhk} classification results. Specifically, $*$ means we employ stronger video-level classification results used in InternVideo~\cite{wang2022internvideo} for a fair comparison. We report our best results in \textbf{bold}, and the previous best results in \underline{underline}, which was achieved by the feature-based method. The last row is achieved when only the last half of backbone layers are adapted; otherwise, full-layer adaptation will lead to out-of-memory on A100-80G.
}
\vspace{-5pt}
\small
\setlength{\tabcolsep}{3pt}
\begin{tabular}{l|ccc|c|ccc>{\columncolor[gray]{0.9}}c|ccccc>{\columncolor[gray]{0.9}}c}
\toprule
\multirow{2}{*}{\textbf{Method}} & \multicolumn{1}{c}{\multirow{2}{*}{\textbf{Backbone}}} & \multicolumn{1}{c}{\multirow{2}{*}{\textbf{E2E}}} & \multirow{2}{*}{\textbf{Flow}} & \multirow{2}{*}{\textbf{Mem}} & \multicolumn{4}{c|}{\textbf{ActivityNet-1.3 }}                                                       & \multicolumn{6}{c}{\textbf{THUMOS14}}                                        \\
\cline{6-15} 
                   &               &            &                      &                      & \textbf{0.5} & \textbf{0.75} & \textbf{0.95} & \textbf{Avg.} & \textbf{0.3} & \textbf{0.4} & \textbf{0.5} & \textbf{0.6} & \textbf{0.7} & \textbf{Avg.} \\
\hline
% TAL-Net~\cite{chao2018rethinking} & I3D &\xmark &\cmark&- &38.23 & 18.30 & 1.30 & 20.22 &53.2 & {48.5} & {42.8} & {33.8} & 20.8 \\ 
BMN~\cite{lin2019bmn}&TSN&\xmark&\cmark& - & {50.07} & {34.78} & {8.29} & {33.85} &{56.0} & 47.4 & 38.8 & 29.7 & 20.5 & 38.5 \\
% {G-TAD}~\cite{xu2020g}  & TSN&\xmark&\cmark&- &{50.36} & {34.60} & {9.02} & {34.09} &{54.5} & {47.6} & {40.2} & {30.8} & {23.4} & 39.3  \\
% TSI~\cite{liu2020tsi} &TSN &\xmark&\cmark&- & 51.18 & 35.02 & 6.59 & 34.15 & 61.0 & 52.1 & 42.6 &  33.2 & 22.4 \\
% BC-GNN~\cite{bai2020boundary} &TSN&\xmark&\cmark&-&50.56 & 34.75 & {9.37} & 34.26 & 57.1 & 49.1 & 40.4 & 31.2 & 23.1 \\
TadTR~\cite{liu2022end}  & I3D &\xmark&\cmark & - & 49.10 & 32.60 & 8.50 & 32.30 & 62.4 & 57.4 & 49.2 & 37.8 & 26.3 & 46.6 \\
% {VSGN~\cite{zhao2021video}} &TSN &\xmark&\cmark& - &{52.38}  & {36.01}  & 8.37 &{35.07} & {66.7} &  {60.4} & {52.4} & {41.0} & {30.4} & 50.2 \\
ActionFormer~\cite{zhang2022actionformer} &SlowFast-R50&\xmark&\xmark&- & 54.26 & 37.04 & 8.13 & 36.02 & 78.7 & 73.3 & 65.2 & 54.6 & 39.7 & 62.3 \\
ActionFormer~\cite{zhang2022actionformer} &I3D&\xmark&\cmark&- &53.50 & 36.20 &8.20 & 35.60 &82.1 & 77.8 & 71.0 &59.4 &43.9 & 66.8\\
% PBRNet~\cite{liu2020progressive} &I3D&\cmark& \cmark&-&{53.96} & 34.97 & 8.98 & 35.01 &58.5 & 54.6 & {51.3} & {41.8}  & {29.5}  \\
ASL~\cite{shao2023action}  & I3D &\xmark &\cmark& - & 54.10 & 67.40 & 8.00 & 36.20 & 83.1 & 79.0 & 71.7 & 59.7 & 45.8 & 67.9  \\
TriDet~\cite{shi2023tridet} & I3D &\xmark &\cmark& - & 54.70 & 38.00 & 8.40 & 36.80 & 83.6 & 80.1 & 72.9 & 62.4 & 47.4 & 69.3 \\
VideoMAEv2~\cite{wang2023videomae}  & VideoMAEv2-g &\xmark &\xmark & - & - & - & - & - & - & - & - & - & - & 69.6 \\
InternVideo~\cite{wang2022internvideo}  & \scriptsize{VideoMAE-H+UniformerV2} &\xmark &\xmark & - & - & - & - & \underline{39.00$^*$} & - & - & - & - & - & \underline{71.5} \\

\hline
\hline
% R-C3D~\cite{xu2017r}  &C3D&\cmark&\xmark& - &26.80 & - & - & -  &44.8 & 35.6 & 28.9 & - & -  \\
% DaoTAD~\cite{wang2021rgb}  & I3D &\cmark & \xmark & 11 & - & - & - & - & 62.8 & - & 53.8 & - & 30.1 \\
AFSD~\cite{lin2021learning} &I3D&\cmark &\cmark&12 &52.40 & 35.30 & 6.50& 34.40 &67.3 &62.4 &55.5 & 43.7 & 31.1 & 52.0 \\
E2E-TAD~\cite{liu2022empirical} & SlowFast-R50 &\cmark &\xmark & 12 & 50.47 & 35.99 & 10.33 & 35.10 & 69.4 & 64.3 & 56.0 & 46.4 & 34.9 & 54.2 \\
BasicTAD~\cite{yang2023basictad} & SlowOnly-R50 &\cmark &\xmark & 12 & 51.20 & 33.41 & 7.57 & 33.12 & 75.5 & 70.8 & 63.5 & 50.9 & 37.4 & 59.6\\
TALLFormer~\cite{cheng2022tallformer}&VideoSwin-B &\cmark &\xmark & 29&54.10 &36.20& 7.90& 35.60 &76.0 &-&63.2&-& 34.5 & 59.2 \\
Re$^2$TAL~\cite{zhao2023re2tal} &Re$^2$VideoSwin-T &\cmark &\xmark & 24 & 54.75 & 37.81 & 9.03 & 36.80 & 77.0 & 71.5 & 62.4 & 49.7 & 36.3 & 59.4 \\
\hline
\textbf{\methodname{}}  & SlowFast-R50 &\cmark &\xmark   & 4.3 & 55.28 & 38.11 & 8.87 & 37.11 & 81.0 & 76.2 & 69.4 & 59.0 & 44.5 & 66.0  \\
\textbf{\methodname{}}  & VideoMAE-S &\cmark &\xmark & 2.5  & 56.15 & 38.99 & 9.07  & 37.85 & 84.5 & 80.2 & 71.6 & 60.9 & 46.9 & 68.8  \\
\textbf{\methodname{}}  & VideoMAE-B &\cmark &\xmark & 4.9  & 56.77 & 39.35 & 9.71  & 38.39 & 87.0 & 82.4 & 75.3 & 63.8 & 49.2 & 71.5  \\
\textbf{\methodname{}}  & VideoMAE-L &\cmark &\xmark & 11.0 & 57.69 & 40.56 & 10.13 & 39.22 & 87.7 & 84.1 & 76.7 & 66.4 & 52.4 & 73.5  \\
\textbf{\methodname{}}  & VideoMAE-H &\cmark &\xmark & 19.2 & 58.04 & 40.55 & 9.75  & 39.37 & 88.9 & 85.3 & 78.6 & 66.9 & 52.5 & 74.4  \\
\textbf{\methodname{}}  & VideoMAEv2-g &\cmark &\xmark & 29.9 & 58.45  & 41.16  &  10.45  & 39.79  & 89.5 & 85.8 & 78.9 & 67.3 & 52.6 & 74.8  \\
\hline
\textbf{\methodname{}$^{\dagger}$} (1536$\times$224$^2$)  & VideoMAEv2-g &\cmark &\xmark & 43.6 & 60.82  & 42.69  &  9.84  & 41.15$^*$  & 89.6 & 85.9 & 79.4 & 67.6 & 53.8 & 75.4  \\
\textbf{\methodname{}} (1536$\times$224$^2$)  & VideoMAEv2-g &\cmark &\xmark & 50.6 & \textbf{61.72}  & \textbf{43.35}  & \textbf{10.85} & \textbf{41.93$^*$}  & \textbf{89.7} & \textbf{86.7} & \textbf{80.9} & \textbf{71.0} & \textbf{56.1} & \textbf{76.9}  \\
\bottomrule
\end{tabular}
\vspace{-8pt}
\label{tab:sota_anet_thumos}
\end{table*}

\subsection{Datasets and Metrics}
We choose ActivityNet-1.3~\cite{caba2015activitynet}, THUMOS14~\cite{jiang2014thumos}, and Epic-Kitchens 100~\cite{damen2018scaling} to evaluate our proposed approach. ActivityNet-1.3 and THUMOS14 are web-collected third-person untrimmed videos, consisting of 19,994 and 413 videos, respectively. EPIC-Kitchens 100 is collected from 700 egocentric videos. Since the action categories of EPIC-Kitchens 100 are more domain-specific and different from common pretraining data, achieving higher performance on this dataset is more challenging. Moreover, we also evaluate our method on the Ego4D-MQ~\cite{grauman2022ego4d} dataset, and the results can be found in the appendix. %~\ref{ego4d_results}.

Following common practice, we report the mean Average Precision (mAP) at certain IoU thresholds and average mAP as the evaluation metrics. On ActivityNet-1.3, the IoU thresholds are chosen from 0.5 to 0.95 with 10 steps. On THUMOS14, the threshold is chosen from \{0.3,0.4,0.5,0.6,0.7\}. On EPIC-Kitchens 100, the threshold is set to \{0.1,0.2,0.3,0.4,0.5\}.

\subsection{Implementation Details}

We implement our method with PyTorch 2.0 and MMAction2~\cite{2020mmaction2} with 4 A100 GPUs. By default, mixed-precision training and activation checkpointing are adopted to save memory. We use ActionFormer~\cite{zhang2022actionformer} as our detection head, and keep the hyper-parameters unchanged on each dataset. The learning rate for the adapter in the backbone is grid-searched from 5e-4 to 5e-5, and other parameters inside the backbone are frozen. On ActivityNet-1.3, we resize the video into a fixed length of 768 frames. On THUMOS14, we randomly truncate a window with 768 frames with a temporal stride of 4. On EPIC-Kitchens 100, we randomly truncate a window with 6144 frames with a temporal stride of 2. After the video encoder, the feature is resized to fixed lengths of 192, 768, and 768, respectively, for the three datasets. Frame resolution is set to 160$^2$ by default. In all experiments, we report the training memory usage. More implementation details can be found in the appendix. %~\ref{implement}.

\subsection{Comparison with SoTA Methods}

\begin{table}[t]
\caption{\textbf{Results on EPIC-Kitchens 100 validation set.} For comparison, the feature-based methods use the same SlowFast-R50. }
\vspace{-5pt}
\label{table:epic}
\small
\setlength{\tabcolsep}{1pt}
\begin{tabular}{lc|ccccc>{\columncolor[gray]{0.9}}c}
\toprule
\textbf{Method} & \textbf{E2E} & \textbf{0.1} & \textbf{0.2} & \textbf{0.3} & \textbf{0.4} & \textbf{0.5} &\textbf{Avg.} \\
\midrule
\textit{Verb Task} \\
\hline
BMN~\cite{lin2019bmn}  & \xmark & 10.8 & 8.8 & 8.4 & 7.1& 5.6 & 8.4 \\
G-TAD~\cite{xu2020g} & \xmark & 12.1 & 11.0 & 9.4 & 8.1 & 6.5 & 9.4 \\
ActionFormer~\cite{zhang2022actionformer} & \xmark & 26.6 & 25.4 & 24.2 & 22.3 & 19.1 & 23.5 \\
ASL~\cite{shao2023action} &\xmark & 27.9 & - & 25.5 & - & 19.8 & 24.6 \\
TriDet~\cite{shi2023tridet} & \xmark &28.6 & 27.4  & 26.1 & 24.2  & 20.8  & 25.4 \\
\methodname{} (SlowFast-R50)   & \cmark  & 26.5 & 25.7 & 23.9 & 21.7 & 17.6 & 23.1 \\
\hline
ActionFormer (VideoMAE-L) &\xmark & 32.7 & 31.6 & 29.1 & 26.7 & 23.6 & 28.7 \\
\textbf{\methodname{} (VideoMAE-L)}   & \cmark  & \textbf{33.1} & \textbf{32.2} & \textbf{30.4} & \textbf{27.5} & \textbf{23.1} & \textbf{29.3} \\
\hline
\hline
\textit{Noun Task} \\
\hline
BMN ~\cite{lin2019bmn} & \xmark & 10.3 & 8.3 & 6.2 & 4.5 & 3.4 & 6.5 \\
G-TAD~\cite{xu2020g} & \xmark & 11.0 & 10.0 & 8.6 & 7.0 & 5.4 & 8.4 \\
ActionFormer~\cite{zhang2022actionformer} & \xmark & 25.2 & 24.1 & 22.7 & 20.5 & 17.0 & 21.9 \\
ASL~\cite{shao2023action} &\xmark & 26.0 & - & 23.4 & - & 17.7 & 22.6 \\
TriDet~\cite{shi2023tridet} & \xmark & 27.4  & 26.3  & 24.6 & 22.2  & 18.3  & 23.8 \\
\methodname{} (SlowFast-R50)   & \cmark  & 24.5 & 23.6 & 22.3 & 20.0 & 16.5 & 21.4 \\
\hline
ActionFormer (VideoMAE-L) &\xmark & 31.3 & 29.7 & 27.2 & 25.3 & 21.3 & 26.9 \\
\textbf{\methodname{} (VideoMAE-L)}  & \cmark   & \textbf{32.4} & \textbf{31.6} & \textbf{30.1} & \textbf{27.4} & \textbf{24.6} & \textbf{29.3} \\
\bottomrule
\end{tabular}
\vspace{-14pt}
\label{tab:sota_epic}
\end{table}

Table \ref{tab:sota_anet_thumos} compares our \methodname{} with other state-of-the-art (SoTA) methods on ActivityNet-1.3 and THUMOS14 datasets. Initially, we use SlowFast-R50 as the backbone. For comparison, we also extract corresponding offline features, utilizing the snippet representation where each snippet comprises 32 frames with $224^2$ resolution. We observe that end-to-end training enhances performance from 62.3\% to 66.0\% on THUMOS14. Notably, this architecture has also been used in E2E-TAD~\cite{liu2022empirical}. However, our method consumes less memory while achieving superior performance. This apple-to-apple comparison underscores the benefits of adapter tuning and the scaling-up principle. 

Furthermore, when adopting the VideoMAE~\cite{tong2022videomae} family as our backbone and progressively scaling up the model size, the performance of  \methodname{} consistently improves. Using the largest model, \ie, VideoMAEv2-giant with 1.01 billion parameters, and larger input data, \ie, 1536 frames with 224$^2$ resolution, we attain an impressive 41.9\% mAP on ActivityNet-1.3 and 76.9\% mAP on THUMOS14. It is noteworthy that the previous SoTA was achieved by VideoMAEv2~\cite{wang2023videomae} and InternVideo~\cite{wang2022internvideo}, which utilize the same detector head as ours but with offline snippet features. Our method surpasses these in detection performance by a large margin, marking the first instance where an end-to-end TAD method can outperform SoTA feature-based results.

We also present our results on EPIC-Kitchens 100 in Table~\ref{tab:sota_epic}. Since videos in this dataset have a longer duration, all previous methods rely solely on pre-extracted features \cite{zhang2022actionformer,tang2023temporalmaxer,shi2023tridet}. Our approach is the first to adopt end-to-end training on this dataset. 
For fair comparison, we first utilize the same backbone as used in previous methods, \ie, SlowFast-R50 pretrained on EPIC, and we achieve comparable performance to ActionFormer~\cite{zhang2022actionformer}. Moreover, when we scale up the backbone to VideoMAE-L~(it is also trained on EPIC-Kitchens 100 classification task), we achieve SoTA performance of 29.3\%. %Further discussions are available in Table~\ref{tab:epic_ablation}.

\subsection{Ablation and Analysis}

In this section, we present a series of analyses to evaluate our proposed method and affirm the benefits of scaling up in TAD. Unless otherwise stated, our experiments utilize a standard input of 768 frames per video on THUMOS14.

\begin{table}[t]
\caption{\textbf{Compared to full fine-tuning, our adapter tuning can achieve better performance with less memory.} Param. is the number of tunable parameters in the backbone. $*$ means out of memory on A100-80GB, and we report the estimated number.}
\vspace{-6pt}
\centering
\small
\setlength{\tabcolsep}{3.5pt}
\begin{tabular}{clcc|c>{\columncolor[gray]{0.9}}c}
    \toprule
    \textbf{Model} &\textbf{Setting}   &\textbf{E2E}  & \textbf{Param.} & \textbf{Mem.}  & \textbf{mAP}\\
    \midrule
    \multirow{4}{*}{VideoMAE-S}   
    & Feature                  &\xmark    & 0   &  - &  57.6 \\
    & Snippet Full FT                &\cmark    & 22M   & 24.6G &  68.4 \\
    & Frame Full FT                 &\cmark    & 22M  & 2.8G &  67.1 \\
    & \textbf{\methodname{}}  &\cmark    & \textbf{1M}  & \textbf{2.5G} &  \textbf{68.8}  \\
    \midrule
    \multirow{4}{*}{VideoMAE-B}   
    & Feature                  &\xmark    & 0   & - &  64.7 \\
    & Snippet Full FT          &\cmark    & 86M   & 87.4G$^*$ &  - \\
    & Frame Full FT            &\cmark    & 86M & 5.6G &  70.1  \\
    & \textbf{\methodname{}}  &\cmark    & \textbf{4M}  & \textbf{4.9G} & \textbf{71.5}  \\ 
    \midrule
    \multirow{4}{*}{VideoMAE-L}   
    & Feature                  &\xmark    & 0   & - &  66.5 \\
    & Snippet Full FT                &\cmark    & 304M   & 193G$^*$ &  - \\
    & Frame Full FT                 &\cmark    & 304M & 13.1G  &  73.0  \\
    & \textbf{\methodname{}}  &\cmark    & \textbf{14M}  & \textbf{11.0G} & \textbf{73.5}  \\ 
    \bottomrule
\end{tabular}
\label{tab:advantage_peft}
\vspace{-12pt}
\end{table}

\vspace{2pt}
\noindent\textbf{The advantage of adapter tuning.} In Table~\ref{tab:advantage_peft}, we compare conventional full fine-tuning with our proposed design. It is evident that end-to-end approaches significantly outperform pre-extracted features. Moreover, with full fine-tuning, the snippet representation slightly advances over frame representation but incurs tremendous memory costs, which aligns with our analysis in Section~\ref{sec:e2e_baseline}. However, \methodname{} uses less memory and still achieves better performance than conventional full fine-tuning. This also verifies the limitations of full fine-tuning, as discussed in Section~\ref{sec:ti_adapter}. Specifically, our method enhances VideoMAE-S backbone with an 11.2\% gain using only 1M trainable parameters. Additionally, Table~\ref{tab:advantage_peft} also demonstrates that scaling up the model size of the video backbone is an effective way to improve TAD performance.

\vspace{2pt}
\noindent\textbf{The advantage of scaling up the data.} 
In addition to model scaling, Table \ref{tab:sale_up_data} verifies the effectiveness of data scaling, which involves two aspects: frame number and frame resolution. Firstly, given the same video duration, increasing the frame number from 768 to 3072 can raise the mAP from 68.8\% to 70.6\%. In the meantime, the memory usage is nearly three times larger. Secondly, increasing the frame resolution from 160$^2$ to 224$^2$ also improves the mAP. In the end, by only scaling up the data, we elevate the mAP from 68.8\% to 71.5\%, already surpassing the current SoTA feature-based approach with a giant backbone model~\cite{wang2022internvideo}.

Moreover, increasing the frame resolution from 160$^2$ to 224$^2$ or increasing the frame number from 768 to 1536 results in the same memory usage of 3.8G. However, the former achieves 70.7\% mAP while the latter only reaches 69.7\%. This suggests that frame resolution may be prioritized under the same memory budget, for the TAD task.

\begin{table}[t]
\caption{\textbf{When scaling up the input data, \methodname{}'s performance consistently increases.} $*$ means snippet representation is used in offline feature extraction, and each snippet has 16 frames.}
\vspace{-7pt}
\centering
\small
\setlength{\tabcolsep}{2.5pt}
\begin{tabular}{llcc|c>{\columncolor[gray]{0.9}}c}
    \toprule
    \textbf{Setting}      & \textbf{Model}  & \textbf{Resolution} & \textbf{Frames}  & \textbf{Mem.}  & \textbf{mAP} \\
    \midrule
    Feature               & VideoMAEv2-g       & 224$^2$ & 768x16$^*$  & - & 69.6 \\
    \midrule
    \multirow{6}{*}{\textbf{\methodname{}}}    &   \multirow{6}{*}{VideoMAE-S}     
                         & 160$^2$ & 768     & 2.5G   & 68.8 \\
                    &    & 160$^2$ & 1536    & 3.8G   & 69.7  \\
                    &    & 160$^2$ & 3072    & 6.5G   & \textbf{70.6}   \\
    \cmidrule{3-6}
                    &    & 224$^2$ & 768     & 3.8G   & 70.7 \\
                    &    & 224$^2$ & 1536    & 6.4G   & 71.0 \\
                    &    & 224$^2$ & 3072    & 11.6G  & \textbf{71.5} \\
    \bottomrule
\end{tabular}
\label{tab:sale_up_data}
\vspace{-8pt}
\end{table}

\begin{table}[t]
\caption{\textbf{\methodname{}$^{\dagger}$ can further push the boundaries of scaling up.} OOM means out of memory on A100-80GB.}
\centering
\vspace{-7pt}
\small
\setlength{\tabcolsep}{2.5pt}
\begin{tabular}{llcc|c>{\columncolor[gray]{0.9}}c}
    \toprule
    \textbf{Setting}      & \textbf{Model}  & \textbf{Resolution} & \textbf{Frame}  & \textbf{Mem.}  & \textbf{mAP} \\
    \midrule
    \multirow{3}{*}{\textbf{\methodname{}}}                  
     & VideoMAE-L    & 160$^2$ & 768     & 11.0G   & 73.5 \\
     & VideoMAEv2-g    & 160$^2$ & 768     & 29.9G   & 74.8 \\
     & VideoMAEv2-g    & 224$^2$ & 1536    & OOM     &  -  \\
    \midrule
    \multirow{3}{*}{\textbf{\methodname{}$^{\dagger}$}} 
    & VideoMAEv2-g & 160$^2$ & 768 & 22.8G  & 73.7 \\
    & VideoMAEv2-g & 224$^2$ & 768 & 30.0G & 74.6 \\
    & VideoMAEv2-g & 224$^2$ & 1536 & 43.6G & \textbf{75.4} \\
    \bottomrule
\end{tabular}
\label{tab:relocate}
\vspace{-14pt}
\end{table}

\vspace{2pt}
\noindent\textbf{The advantage of \methodname{}$^\dagger$.} 
Given the effectiveness of scaling up the model or data, we further explore combining these approaches. In Table~\ref{tab:relocate}, using 768 frames while scaling up the model to VideoMAEv2-giant results in memory usage escalating to 29.9G. In such a scenario, further increasing the data could easily lead to memory overflow, even with the A100-80GB GPU. This indicates that adaptation tuning reached its limit under this extreme case. Therefore, to utilize both the largest models with larger data simultaneously, \methodname{}$^\dagger$ shows its advantage. 

Concretely, when switching from \methodname{} to \methodname{}$^{\dagger}$, the memory usage of the VideoMAEv2-giant model is reduced from 29.9G to 22.8G. Although a slight performance drop is observed, its reduced memory footprint enables scaling up data from 768 frames to as much as 1536 frames with a high resolution of 224$^2$.
This scalability helps mitigate the performance drop and achieves a higher mAP of 75.4\%.

% \subsubsection{The ablation of different adapter design}
\vspace{2pt}
\noindent\textbf{The ablation of the adapter design.} 
Detailed in Table~\ref{tab:ablation_adapter}, we compare different adapter architectural designs. The baseline, \ie, offline snippet feature, achieves 64.7\% mAP. End-to-end learning in all designs yields at least a 5\% improvement. Our \methodname{} achieves 71.5\% in the end. Compared to standard adapter~\cite{houlsby2019parameter}, ours consumes similar memory but achieves higher mAP. This verifies that local temporal context is vital for the TAD task. In contrast to full fine-tuning (FT), we tune only 4M parameters using less memory. In our design, we find that removing the residual connection of the depth-wise convolution drops performance by 0.7\%, and training becomes unstable. We also implement an adaptation design proposed in LongLoRA~\cite{chen2023longlora}, which efficiently computes long-range attention and shows decent performance but requires more parameters and memory.

\begin{table}[h]
\caption{\textbf{Ablation of different adapter architectural designs.} VideoMAE-B is used to conduct the following experiments.}
\centering
\vspace{-8pt}
\small
\setlength{\tabcolsep}{2.5pt}
\begin{tabular}{lc|cc>{\columncolor[gray]{0.9}}cc}
    \toprule
    \textbf{Setting}   &\textbf{E2E}  & \textbf{Param.} & \textbf{Mem.}  & \textbf{mAP} & \textbf{gains}\\
    \midrule
    Snippet Feature                  &\xmark    & 0   & - &  64.7\\
    \midrule
    + Full FT                 &\cmark    & 86M   & 5.6G &  70.1 & + 5.1 \\
    + LongLoRA~\cite{chen2023longlora}      &\cmark    & 28M   & 6.2G &  71.1 & + 6.1 \\    
    + Standard Adapter~\cite{houlsby2019parameter}           &\cmark    & 3.6M  & 4.8G &  70.2 & + 5.2 \\
    + \methodname{} (w/o residual )   &\cmark    & 4.0M  & 4.9G &  70.8 & + 5.8 \\
    + \textbf{\methodname{}}                    &\cmark    & \textbf{4.0M}  & \textbf{4.9G} &  \textbf{71.5} & \textbf{+ 6.5} \\
    \bottomrule
\end{tabular}
\label{tab:ablation_adapter}
\vspace{-8pt}
\end{table}

\vspace{2pt}
\noindent\textbf{The necessity of end-to-end training for TAD.} 
As previously discussed, end-to-end training can address discrepancies between the pretraining and fine-tuning stages in terms of training data and learning tasks. To corroborate this, we employ models pretrained on different datasets for the EPIC-Kitchens TAD task in Table~\ref{tab:epic_ablation}. Kinetics-400 (K400)~\cite{kay2017kinetics} represents commonly collected third-person web data and exhibits a large domain gap compared to EPIC-Kitchens 100. Using K400 for pretraining, we observe that end-to-end TAD training allows for +5.56 gain.
Conversely, using a model already finetuned on EPIC-Kitchens still yields a +2.32 improvement. Unlike K400 pretraining, since this model has already adapted to the data discrepancy, we can infer that this gain leverages differences between the classification task in pretraining and the detection task in fine-tuning. Such improvements further underscore the significance of end-to-end training in TAD.

\begin{table}[t]
\caption{\textbf{End-to-end TAD can alleviate the discrepancies between pretraining and finetuning.} VideoMAE-L with different pretrained weights are used on the EPIC-Kitchens 100 Noun task.}
\centering
\vspace{-6pt}
\small
\setlength{\tabcolsep}{2.5pt}
\begin{tabular}{lc|ccc>{\columncolor[gray]{0.9}}cc}
    \toprule
    \textbf{Pretrain Dataset}      & \textbf{E2E}  & \textbf{0.1} & \textbf{0.3} & \textbf{0.5}  & \textbf{mAP} & \textbf{gain} \\
    \midrule
    K400~\cite{kay2017kinetics} & \xmark & 18.69 & 16.35 & 11.52 & 15.77  \\
    K400~\cite{kay2017kinetics}& \cmark & 24.33 & 22.14 & 16.87 & \textbf{21.33} & \textbf{+ 5.56} \\
    \midrule    
    K400~\cite{kay2017kinetics} + EPIC~\cite{damen2018scaling}  & \xmark &  31.32 & 27.25 & 21.33 & 26.98  \\
    K400~\cite{kay2017kinetics} + EPIC~\cite{damen2018scaling}  & \cmark & 32.41 & 30.13 & 24.59 & \textbf{29.30} & \textbf{+ 2.32}\\
    \bottomrule
\end{tabular}
\vspace{-16pt}
\label{tab:epic_ablation}
\end{table}

\subsection{Error Analyses}
\begin{figure}[h]
\centering
\includegraphics[width=1.0\linewidth]{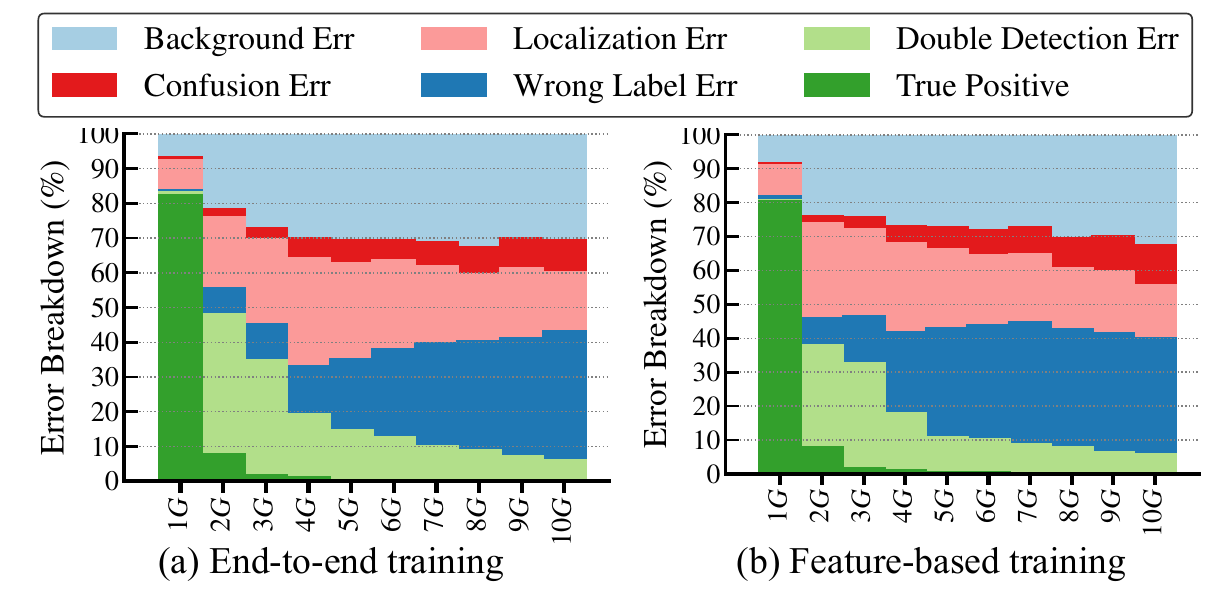}
\vspace{-19pt}
\caption{\textbf{False Positive Profiling on THUMOS14 using~\cite{detad}}. We use VideoMAEv2-giant as the backbone, and compare end-to-end training with pre-extracted feature-based training.}
\label{fig:visual}
\vspace{-18pt}
\end{figure}

We also conduct false positive analysis at tIoU=0.5 in Fig.~\ref{fig:visual}. Compared to feature-based training, learning from raw frames produces more helpful positive detections. More importantly, the percentage of wrong label error is reduced after end-to-end training, suggesting its unique advantage in classifying accurate action labels. 

\vspace{-10pt}

\section{Conclusions}

This work introduces a memory-efficient and parameter-efficient end-to-end method named \textbf{\methodname}. Our key innovation lies in the proposed temporal-informative adapter, which is tailored for TAD with low computation costs. Furthermore, we design an alternative placement for adapters to minimize memory usage. By demonstrating the feasibility and effectiveness of scaling up end-to-end TAD, our work achieves new SoTA performance across multiple datasets. Particularly, this is the first instance of an end-to-end TAD method that surpasses the current best feature-based models. In fact, AdaTAD achieves a groundbreaking 75.4\% mAP on THUMOS14. We believe our work underscores the possible paradigm shift in TAD, advocating a move away from the traditional methodology of separate feature extraction and offline detection towards a more integrated approach of scaling up end-to-end TAD training.

\noindent \textbf{Acknowledgement.} This work was supported by the King Abdullah University of Science and Technology (KAUST) Office of Sponsored Research through the Visual Computing Center (VCC) funding, as well as the SDAIA-KAUST Center of Excellence in Data Science and Artificial Intelligence (SDAIA-KAUST AI).

\section{Appendix}
\renewcommand\thesection{\Alph{section}}
\renewcommand\thesubsection{\thesection.\arabic{subsection}}
\setcounter{section}{0}
In this appendix, we provide more details of our method and present more experiment results. Specifically, we give a detailed illustration of \methodname$^\dagger$ in Section~\ref{adatad+}. Then, we present the implementation details between different datasets in Section~\ref{implement}. Next, we provide our results on the Ego4D dataset in Section~\ref{ego4d_results}. After this, we show additional experiments and further analysis in Section~\ref{experiments}. Then, the error analysis is conducted in Section~\ref{error}, and qualitative visualization is demonstrated in Section~\ref{visual}. Finally, we discuss the limitations of our work in Section~\ref{limitations}.

\section{Further illustration of \methodname$^\dagger$}
\label{adatad+}

\begin{figure}[t]
\centering
\includegraphics[width=0.75\linewidth]{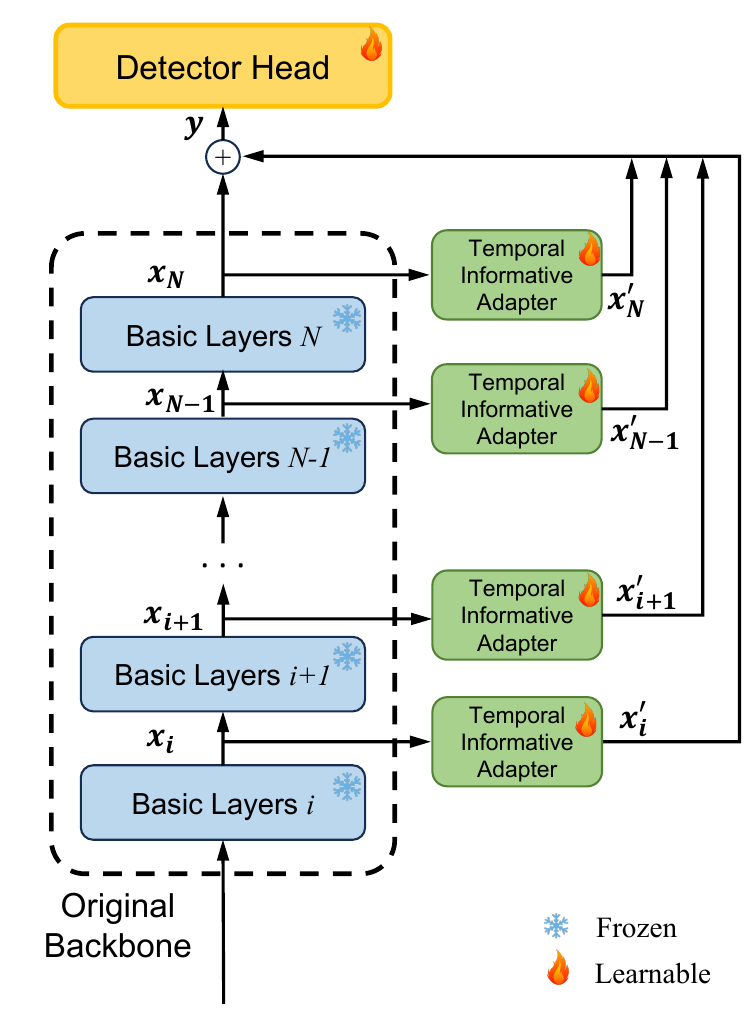}
\caption{\textbf{Detailed architecture of \methodname{}$^\dagger$}. The output of each adapter will be added to the final output of the original backbone.}
\label{fig:framework_adatad_plus}
\vspace{-8pt}
\end{figure}

To reduce the memory usage and further scale up the model and data, \methodname$^\dagger$ proposes an alternative placement for the temporal-informative adapters (TIAs). The entire pipeline of \methodname{}$^\dagger$ is shown in Fig.~\ref{fig:framework_adatad_plus}. 

Concretely speaking, we retain the TIA architecture, but eliminate the last residual connection in \methodname{}, as illustrated in Equation~\ref{eq:ada_plus_adapter}. Therefore, given layer $\bm i$ 's output $\bm{x_i}$, the corresponding TIA's output $\bm{x'_i}$ will be 0 at the start of the training, since the weights and biases of $\bm W_{\rm up}$ are initialized to 0.
\begin{equation}
\begin{aligned}
 \bm{x'_i} &= \bm {\sigma} ({\bm W_{\rm down}^{\rm \top}} \cdot \bm{x_i} ), \\
 \bm{x'_i} &= {\bm W_{mid}^{\rm \top}} \cdot \operatorname{DWConv}_k(\bm {x'_i}) + \bm {x'_i}, \\
 \bm{x'_i} &= \alpha \cdot {\bm W_{\rm up}^{\rm \top}} \cdot \bm{x'_i}.
\end{aligned}
\label{eq:ada_plus_adapter}
\end{equation}

What's more, in the above equation, $\bm{x'_i}$ is not added into $\bm{x_i}$, which is different from \methodname{}. This means that the adapter's outputs do not contribute to the middle activations of the original backbone. Instead, it serves as a \textit{residual} and is directly added to the backbone's final output $\bm{x_N}$, where $N$ is the total number of layers in the backbone. Consequently, the output of the video encoder becomes $\bm{y}$, as depicted in Equation~\ref{eq:final_output}. At the beginning of the fine-tuning, $\bm{y}$ is initialized as $\bm{x_N}$, and is gradually augmented with the aforementioned \textit{residuals} to adapt the fine-tuning, which is driven by the updated TIAs.
\begin{equation}
\begin{aligned}
\bm{y} &= \bm{x_N} + \sum_{i=1}^{N} \bm{x'_i}.
\end{aligned}
\label{eq:final_output}
\end{equation}

\begin{table*}[ht]
\caption{\textbf{End-to-End setting in different TAD datasets.}}
\vspace{-6pt}
\label{table:hyper_para}
\centering
\small
\begin{tabular}{lcccc}
\toprule
 \textbf{Config} & \textbf{ActivityNet-1.3~\cite{caba2015activitynet}} & \textbf{THUMOS14~\cite{jiang2014thumos}} & \textbf{EPIC-Kitchens 100~\cite{damen2018scaling}} & \textbf{Ego4D-MQ~\cite{grauman2022ego4d}} \\
\midrule
\textit{\textbf{Backbone Setting}} \\
\midrule
Video Preprocessing & Resize & Sliding Window & Sliding Window & Padding \\
Frame Stride & - & 4 & 2 & 2 \\
Frame Number & 192$\times$4 & 768 & 768$\times$8 & 900$\times$8 \\
Frame Resolution & \multicolumn{4}{c}{160$\times$160}  \\
Data Augmentation & \multicolumn{4}{c}{RandomResizedCrop + Flip + ImgAug + ColorJitter} \\ 
Feature Post Processing & \multicolumn{4}{c}{Spatial Average Pooling + Resize} \\
Feature Resize Length & 192 & 768 & 768 & 900 \\
\midrule
\textit{\textbf{Detector Setting}} \\
\midrule
Warmup Epoch & 5 & 5 & 5 & 5 \\
Total Epoch & 15 & 60 & 35 & 15 \\
Batch Size & 16 & 2 & 2 & 2 \\
\bottomrule
\end{tabular}
\vspace{-6pt}
\end{table*}

In our design, the key to memory reduction lies in \textbf{stopping the gradient backpropagation for the original backbone}. Since the TIA's output directly goes to the final output, gradients during backpropagation do not trace back to the original backbone but are confined to the shallow and lightweight adapters. This architecture can be viewed as the \textit{Side Network}~\cite{sung2022lst} or \textit{Ladder Network}~\cite{yin2023parameter}, meaning that it creates a light network apart from the original heavy backbone. Compared to the traditional PEFT approach or our \methodname{}, this design requires only a few activations from the backbone, and it can further refine these activations to adapt the downstream task during transfer learning. 

\section{Implementation Details}
\label{implement}

\noindent \textbf{Training Details.} Unless otherwise specified, we prefer \methodname{} for its high performance. By default, mixed precision training~\cite{micikevicius2017mixed} and activation checkpointing~\cite{chen2016training} are adopted following previous work~\cite{cheng2022tallformer}. We also utilize flash attention~\cite{dao2022flashattention} to accelerate the computation and save the memory in the VideoMAE-family models. Our method is evaluated on four datasets, with the hyper-parameters detailed in Table~\ref{table:hyper_para}. 

\vspace{4pt}
\noindent \textbf{Data Preprocessing.} Regarding the data preprocessing, we provide the following instructions. For ActivityNet, due to the variable action duration, we resize videos into fixed lengths with 768 frames. After the video encoder, average pooling is adopted along the spatial dimension, and the feature's temporal length is resized to 192. For THUMOS14 and EPIC-Kitchens, given their longer video lengths, we apply the random truncation to a fixed-length window during training and use sliding window during testing. For THUMOS14, the window length is 768 frames with a stride of 4. For EPIC-Kitchens, it's 768$\times$8 frames with a stride of 2, and additionally, the feature after the video encoder is temporally resized to 768. On the Ego4D-MQ dataset, since all the videos are under 8 minutes, we sample 900$\times$8 frames with a stride of 2, and resize the feature to length 900.

\vspace{4pt}
\noindent \textbf{Video Model.} All the action recognition models used in our paper, such as SlowFast~\cite{slowfast_iccv19}, VideoSwin~\cite{liu2022video}, and VideoMAE~\cite{tong2022videomae}, are pretrained on Kinetics-400~\cite{kay2017kinetics}, except the VideoMAEv2-giant~\cite{wang2023videomae} that is hybrid pretrained and fine-tuned on Kinetics-710~\cite{carreira2019short}. Meanwhile, for the ViT-based model, to avoid excessive temporal attention for the untrimmed video, we chunk the video into shorter clips before processing it in the ViT block. For example, given the input video with 768 frames, we reshape the 768 frames as 16$\times48$, dividing it into 48 snippets with 16 frames each. Then VideoMAE processes these snippets independently, but before sending them into our proposed adapter, we reassemble the snippets into the original 768 frames to achieve cross-snippet information exchange.  This approach, focusing temporal attention only on 16 frames, is specific to ViT-based models. CNN-based models like SlowFast and local attention-based models like VideoSwin do not require this reshaping. 

\vspace{4pt}
\noindent\textbf{Memory Usage on EPIC-Kitchens and Ego4D-MQ.} On both datasets, we use the VideoMAE-Large as the backbone, but differently pretrained on respective datasets. On EPIC-Kitchens, the memory usage is 48.5GB per video for $768\!\times\!8\!=\!6144$ frames. On Ego4D-MQ, the memory usage is 60.0GB per video for $900\!\times\!8\!=\!7200$ frames.

\begin{table*}[ht]
\caption{\textbf{Results on the \textit{validation} set of Ego4D-Moment Queries v2.0.} We report \textit{m}AP at different tIoU thresholds. InternVideo~\cite{chen2022internvideo} denotes the backbone is VideoMAE-L~\cite{tong2022videomae}, which is pretrained and fine-tuned on Ego4D-Moment Queries.}
\label{table:ego4d_val}
\centering
\small
\begin{tabular}{llc|ccccc|>{\columncolor[gray]{0.9}}c}
\toprule
 \textbf{Method} & \textbf{Feature} & \textbf{E2E} & \textbf{0.1} & \textbf{0.2} & \textbf{0.3}  & \textbf{0.4} & \textbf{0.5} &\textbf{Avg.}\\
\midrule
VSGN~\cite{zhao2021video}          & EgoVLP      & \xmark & 16.63 & - & 11.45  & - & 6.57  & 11.39  \\
VSGN~\cite{zhao2021video}          & InternVideo & \xmark & -& -& -& - & -  & 19.35  \\
ActionFormer~\cite{zhang2022actionformer}  & EgoVLP & \xmark & 26.84 & -& 20.57 & - & 14.54 & 20.60 \\
ActionFormer~\cite{zhang2022actionformer}  & InternVideo & \xmark & -& -& - & -& - & 23.29 \\
ASL~\cite{shao2023action}           & EgoVLP      & \xmark & 29.45 & - & 23.03 & - & 16.08 & 22.83 \\
\midrule
\methodname{} & InternVideo & \xmark & 32.40 & 29.50 & 26.98 & 24.43 & 21.88 & 27.07 \\
\textbf{\methodname{}} & \textbf{InternVideo} & \textbf{\cmark} & \textbf{33.53} & \textbf{30.71} & \textbf{28.04} & \textbf{25.51} & \textbf{22.59} & \textbf{28.08}\\
\bottomrule
\end{tabular}
\end{table*}

\section{Results on Ego4D-Moment Queries}
\label{ego4d_results}

We present our results on the Ego4D-MQ dataset in Table~\ref{table:ego4d_val}. The backbone is VideoMAE-Large, which is fine-tuned by InternVideo on Ego4D classification task~\cite{chen2022internvideo}. Note that previous methods like VSGN~\cite{zhao2021video} and ActionFormer~\cite{zhang2022actionformer} use the same backbone, yet they extract high-resolution, densely sampled offline features. Our work first establishes a stronger baseline by adopting an improved version of ActionFormer on this dataset~\cite{sui2023nms}, achieving 27.07\% mAP. Furthermore, utilizing end-to-end training with our proposed temporal-informative adapter elevates the performance to 28.08\% mAP.

\textbf{More importantly, when we apply the full fine-tuning on Ego4d-MQ, no performance gain was observed (27.01\% mAP).}  This implies that the pretrained model is sufficiently robust, and the downstream data is too limited for effective transfer learning. Nonetheless, our \methodname{} manages to yield a performance increase of +1.01\%. This outcome further demonstrates the efficacy of our adapter-based transfer learning.

\section{Additional Experiments}
\label{experiments}

In this section, we provide additional experiments to study the effectiveness of our proposed method.  These experiments are omitted from the main paper due to lack of space.

\subsection{More Results of \methodname{}$^\dagger$}

In our paper, we propose two distinct adapter placement designs: \methodname{} and \methodname{}$^\dagger$. The former directly inserts the adapters into the original backbone, while the latter positions the adapter outside the backbone. This external placement in \methodname{}$^\dagger$ stops gradient backpropagation to the original backbone, negating the need to save massive intermediate activations and thus reducing memory usage. However, previous research~\cite{sung2022lst} suggests that such a design might comprise the performance, as the lighter adapters may limit the representation capacity during transfer learning. Therefore, to verify this hypothesis and explore the advantages of \methodname{}$^\dagger$, we conduct experiments summarized in Table~\ref{tab:relocate_more}, leading to two key conclusions:

\vspace{4pt}
\begin{enumerate}
\item \textbf{With identical input data, \methodname{}$^\dagger$ underperforms to \methodname{}.} For example, when using VideoMAE-Base with 768 frames and a resolution of 160$^2$, the performance of \methodname{}$^\dagger$ drops from 71.5\% to 70.2\%. This verifies that the transfer learning ability of \methodname{} is better than  \methodname{}$^\dagger$.
\item \textbf{Under a similar memory budget, \methodname{}$^\dagger$ can achieve comparable performance than \methodname{} by scaling up the input data.} Owing to its lower memory usage, \methodname{}$^\dagger$ allows for larger input data, thereby improving performance. For instance, \methodname{}$^\dagger$ with inputs of 768, 224$^2$ can marginally outperform \methodname{} with inputs of 768, 160$^2$ under the same memory budget.
\end{enumerate}
\vspace{4pt}

These experiments underscore the significance of data scaling. Particularly with larger models like VideoMAEv2-giant, \methodname{} has reached a limit of scaling up the input data. In such scenarios, only \methodname{}$^\dagger$ can manage a similar memory budget while enhancing input data for superior performance. In conclusion, \methodname{}$^\dagger$ is tailored for higher detection performance in situations where the backbone model is very large and cannot accommodate more frames or higher image resolution. \textbf{In scenarios where these are viable, \methodname{} remains the recommended choice for optimal performance.} Therefore, except in the case of giant models, we default to using \methodname{}.

\begin{table}[t]
\caption{\textbf{The advantage of \methodname{}$^{\dagger}$ lies in scaling up the data with low memory usage.} When using the same input, the performance of \methodname{}$^\dagger$ is inferior to \methodname{}. However, under a similar memory budget, \methodname{}$^\dagger$ can achieve comparable or better performance thanks to data scaling. We report the memory usage and average mAP on the THUMOS14 dataset. }
\centering
\small
\begin{tabular}{cll|c>{\columncolor[gray]{0.9}}c}
\toprule
\textbf{Model}   & \textbf{Method} & \textbf{Input}   &  \textbf{Mem.}  & \textbf{mAP} \\
\midrule
\multirow{3}{*}{\textbf{VideoMAE-S}}    
& \methodname{} & 768, 160$^2$ & 2.5G  & 68.8 \\
& \methodname{}$^\dagger$ & 768, 160$^2$ &  1.8G   & 68.0 \\
& \methodname{}$^\dagger$ & 768, 224$^2$ &  \textbf{2.7G}   & \textbf{68.9} \\
\midrule
\multirow{3}{*}{\textbf{VideoMAE-B}}    
& \methodname{} & 768, 160$^2$ & 4.9G  & 71.5 \\
& \methodname{}$^\dagger$ & 768, 160$^2$ &  4.0G   & 70.2 \\
& \methodname{}$^\dagger$ & 768, 224$^2$ &  \textbf{4.9G}   & \textbf{71.9} \\
\midrule
\multirow{3}{*}{\textbf{VideoMAE-L}}    
& \methodname{} & 768, 160$^2$ & 11.0G   & 73.5 \\
& \methodname{}$^\dagger$ & 768, 160$^2$ &  8.1G   & 73.1 \\
& \methodname{}$^\dagger$ & 768, 224$^2$ &  \textbf{10.8G}  & \textbf{73.7} \\
\bottomrule
\end{tabular}
\label{tab:relocate_more}
\end{table}

\subsection{More Results with Swin and SlowFast}

To validate the efficacy of our adapter tuning approach, we expand our study to include a broader range of backbone models. This extended analysis, detailed in Table~\ref{tab:advantage_peft_more}, encompasses not only the window-based transformer model, \ie, VideoSwin~\cite{liu2022video}, but also the 3D CNN model, \ie, SlowFast~\cite{slowfast_iccv19}. The findings are in line with those reported in the main paper, indicating that adapter tuning can yield better detection performance compared to full fine-tuning. Notably, the performance gains are even more pronounced with VideoSwin and SlowFast than with VideoMAE. Specifically, our method enhanced detection performance from 55.1\% to 63.7\% with VideoSwin-B, and from 62.3\% to 66.1\% with SlowFast-R50.

\begin{table}[t]
\caption{\textbf{Compared to full fine-tuning, our adapter tuning can achieve better performance with less memory.} Param. is the number of tunable parameters in the backbone. $*$ means out of memory on A100-80GB, and we report the estimated number. We conduct the following experiments on THUMOS14 dataset.}
\centering
\small
\setlength{\tabcolsep}{3.2pt}
\begin{tabular}{clcc|c>{\columncolor[gray]{0.9}}c}
    \toprule
    \textbf{Model} &\textbf{Setting}   &\textbf{E2E}  & \textbf{Param.} & \textbf{Mem.}  & \textbf{mAP}\\
    \midrule
    \multirow{4}{*}{VideoSwin-B}   
    & Feature                  &\xmark    & 0   &  - &  55.1 \\
    & Snippet Full FT          &\cmark    & 87.6M   & 213G$^*$ & - \\
    & Frame Full FT            &\cmark    & 87.6M  & 16.4G  & 60.4 \\
    & \textbf{\methodname{}}  &\cmark    & 3.9M  & 16.1G  & \textbf{63.7}    \\
    \midrule
    \multirow{4}{*}{SlowFast-R50}   
    & Feature                  &\xmark    & 0   & - & 62.3   \\
    & Snippet Full FT          &\cmark    & 33.6M   & 36.9G &  66.1  \\
    & Frame Full FT            &\cmark    & 33.6M & 3.9G &   64.3 \\
    & \textbf{\methodname{}}  &\cmark    & 11.4M  & 4.3G  &  \textbf{66.0}  \\ 
    \bottomrule
\end{tabular}
\label{tab:advantage_peft_more}
\end{table}

\subsection{Study of Different Kernel Size in TIA}

In our temporal-informative adapter (TIA), we employ depth-wise convolution along the temporal dimension to capture context from adjacent frames. This design utilizes a 3D depth-wise convolution layer with kernel size $(t, h, w)$. By default, the kernel is set to $(3,1,1)$. To assess the influence of various kernel sizes in TIA, we conduct a study summarized in Table~\ref{tab:ablation_kernel_size}.

First, we expand the kernel size to $(3,3,3)$, and note a decrease in performance. This suggests that the spatial context has been effectively handled by the original backbone, and additional spatial processing could potentially disrupt the pretrained knowledge. Subsequently, reducing the kernel size to $(1,1,1)$ results in inferior performance compared to ours, likely due to insufficient temporal information aggregation.  Moreover, we gradually increase the temporal kernel size from 3 to 7, 13, and 21, observing a consistent downward trend in performance. This indicates that aggregating a longer-range temporal context does not necessarily benefit the backbone. Overall, our default TIA configuration demonstrates the best performance.

\begin{table}[ht]
\caption{\textbf{Ablation of different kernel size in depth-wise convlution in \methodname.} The order of the kernel size follows $t, h, w$. VideoMAE-B is used as the backbone on THUMOS dataset.}
\centering
% \vspace{-6pt}
\small
\setlength{\tabcolsep}{11pt}
\begin{tabular}{l|ccc>{\columncolor[gray]{0.9}}c}
    \toprule
    \textbf{DW Kernel}  & \textbf{0.3} & \textbf{0.5} & \textbf{0.7 } & \textbf{mAP} \\
    \midrule
    (3,1,1)        & 87.04 & 75.33 & 49.22 & \textbf{71.56} \\
    \midrule
    (1,1,1)        & 85.97 & 74.61 & 49.12 & 71.06 \\
    \midrule
    (3,3,3)        & 85.46 & 73.74 & 49.13 & 70.63 \\
    \midrule
    (7,1,1)        & 86.17 & 74.62 & 48.93 & 71.02 \\
    (13,1,1)       & 85.68 & 72.80 & 47.98 & 69.96 \\
    (21,1,1)       & 84.75 & 72.53 & 46.74 & 69.03 \\
    \bottomrule
\end{tabular}
\label{tab:ablation_kernel_size}
\end{table}

\subsection{Ablation of Adapter Design on ActivityNet-1.3}

In the main paper, we present a comparison of various adapter architectural designs on THUMOS14. Extending our analysis, we conduct a similar ablation study on ActivityNet-1.3, with results detailed in Table~\ref{tab:ablation_adapter_anet}. The findings from this study align with the conclusions drawn in the main paper. Notably, since the size of ActivityNet is much larger than THUMOS14, the competition in performance metrics is more intense, resulting in smaller gains. Despite this, our \methodname{} still stands out among other adapter architectural designs. Compared to using frozen features, \methodname{} enhances performance from 36.64\% to 38.39\%, affirming its efficacy in a more challenging dataset.

Moreover, we also combine our proposed TIA modules with full fine-tuning strategy, but observe a decreased mAP. This is due to the learning rate conflicts between the pretrained backbone and the newly added adapter. The former prefers a smaller learning rate since the model is pretrained on large datasets, while the latter prefers a larger learning rate since it is newly added and randomly initialized.

\begin{table}[h]
\caption{\textbf{Ablation of different adapter architectural designs on ActivityNet-1.3.} VideoMAE-B is used as the backbone.}
\centering
\small
\setlength{\tabcolsep}{2.5pt}
\begin{tabular}{lc|cc>{\columncolor[gray]{0.9}}cc}
    \toprule
    \textbf{Setting}   &\textbf{E2E}  & \textbf{Param.} & \textbf{Mem.}  & \textbf{mAP} & \textbf{gains}\\
    \midrule
    Snippet Feature                  &\xmark    & -   & - & 36.64  \\
    \midrule
    + LongLoRA~\cite{chen2023longlora}      &\cmark    & 28M   & 6.2G & 37.69  & +1.05  \\    
    + Full FT                 &\cmark    & 86M   & 5.6G & 38.13  & +1.49 \\
    + Plain Adapter~\cite{houlsby2019parameter}           &\cmark    & 3.6M  & 4.8G & 38.21  & +1.57  \\
    + \methodname{} (w.o. residual )   &\cmark    & 4.0M  & 4.9G &  38.23 & +1.59  \\
    + \textbf{\methodname{}}                    &\cmark    & \textbf{4.0M}  & \textbf{4.9G} &  \textbf{38.39} & \textbf{+1.75} \\
    \midrule
    + Full FT + TIA &\cmark    & 90M  & 5.8G &  37.89 & +1.25  \\
    \bottomrule
\end{tabular}
\label{tab:ablation_adapter_anet}
\end{table}

\subsection{\methodname{} with Different Detector Head}

Our end-to-end framework with the introduced adapter is effective not only on ActionFormer~\cite{zhang2022actionformer}, but also with other TAD heads. As shown in Table~\ref{tab:ablation_detector}, we successfully employ the proposed method on GTAD~\cite{xu2020g} and TriDet~\cite{shi2023tridet}. When using VideoMAE-Large as the backbone, we observe significant improvements. For instance, compared to using densely extracted snippet features (16 frames per snippet with a resolution of 224$^2$), our approach elevates GTAD’s performance from 50.8\% to 55.5\%, and TriDet's performance from 68.8\% to 74.1\%. Additionally, compared to ActionFormer, TriDet achieves better detection performance with offline features and end-to-end training, thanks to the proposed  SGP layer and Trident-Head. Overall, our proposed end-to-end training paradigm is agnostic to different action detectors, and it can boost the detector's performance by a large margin.

\begin{table}[ht]
\caption{\textbf{Ablation of different detector heads.} VideoMAE-L is used as the backbone on THUMOS dataset.}
\centering
% \vspace{-6pt}
\small
\setlength{\tabcolsep}{5pt}
\begin{tabular}{ll|ccc>{\columncolor[gray]{0.9}}c}
\toprule
\textbf{Detector Head}   &\textbf{Setting}  & \textbf{0.3} & \textbf{0.5}  & \textbf{0.7} & \textbf{mAP} \\
\midrule
\multirow{2}{*}{GTAD~\cite{xu2020g}}   
& Feature        &  65.8  & 53.6  & 31.3 & 50.8  \\
& \textbf{\methodname{}}  &  \textbf{69.5} & \textbf{57.6}  & \textbf{37.4} & \textbf{55.5} \\
\midrule\multirow{2}{*}{ActionFormer~\cite{zhang2022actionformer}}   
& Feature                 &  82.9  & 70.8  & 42.7 & 66.5 \\
& \textbf{\methodname{}}  &  \textbf{87.7}  & \textbf{76.7}  & \textbf{52.4} & \textbf{73.5} \\
\midrule
\multirow{2}{*}{TriDet~\cite{shi2023tridet}}   
& Feature        & 84.0   & 73.4  & 45.1 & 68.8  \\
& \textbf{\methodname{}}  & \textbf{88.7}  & \textbf{78.1} & \textbf{52.2} & \textbf{74.1}  \\
\bottomrule
\end{tabular}
\label{tab:ablation_detector}
\end{table}

\begin{figure*}[t]
\centering
{\includegraphics[width=0.50\linewidth]{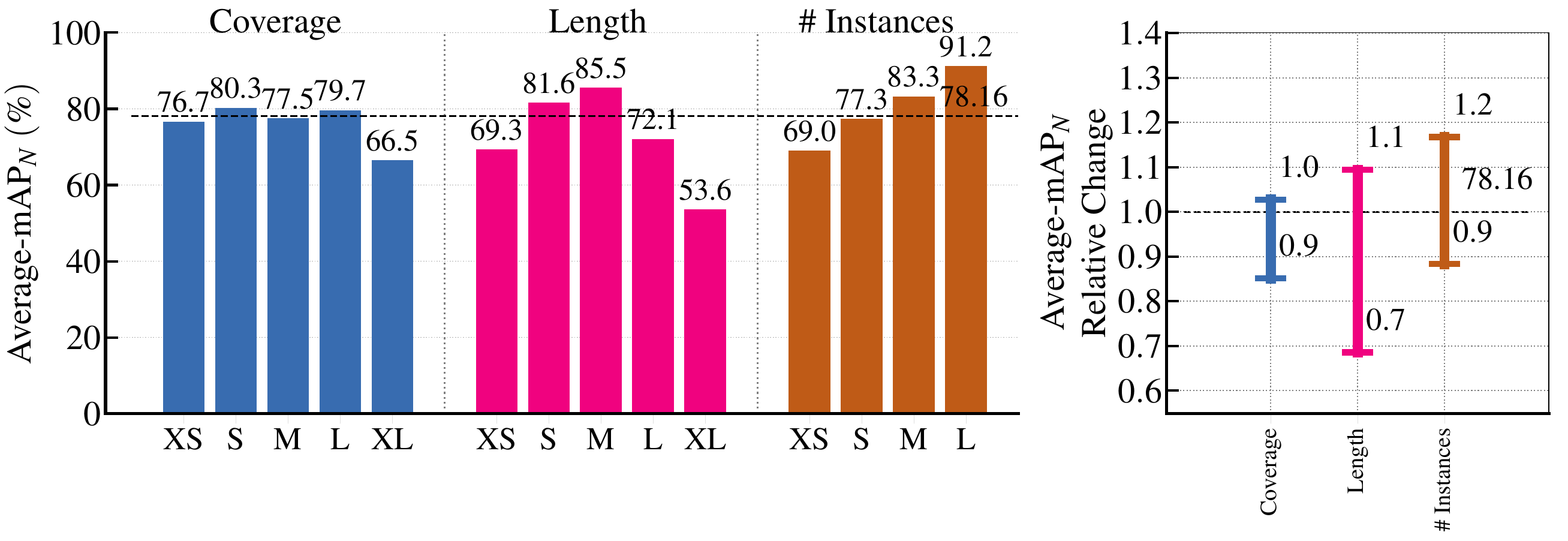} }
\qquad
{\includegraphics[width=0.44\linewidth]{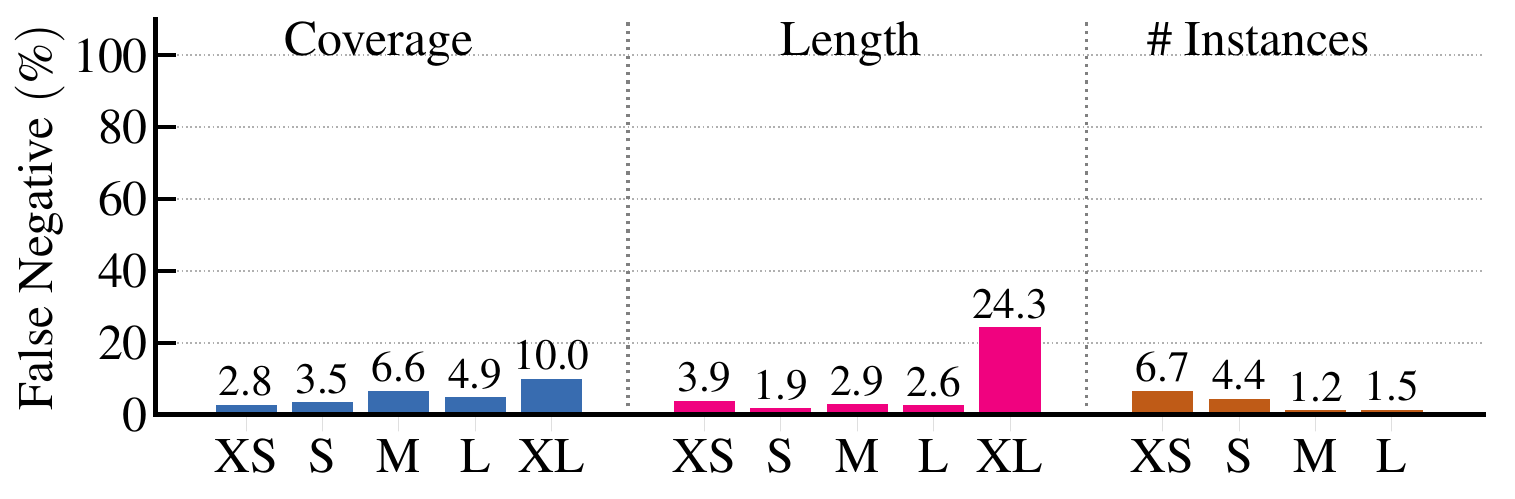} }
\subfloat[\textbf{Sensitive Analysis.} \textit{Left}: normalized mAP at tIoU=0.5 under different video contents. \textit{Right}: The relative normalized mAP change at tIoU=0.5 with respect to different characteristics of the ground truth instances.]{{\includegraphics[width=0.50\linewidth]{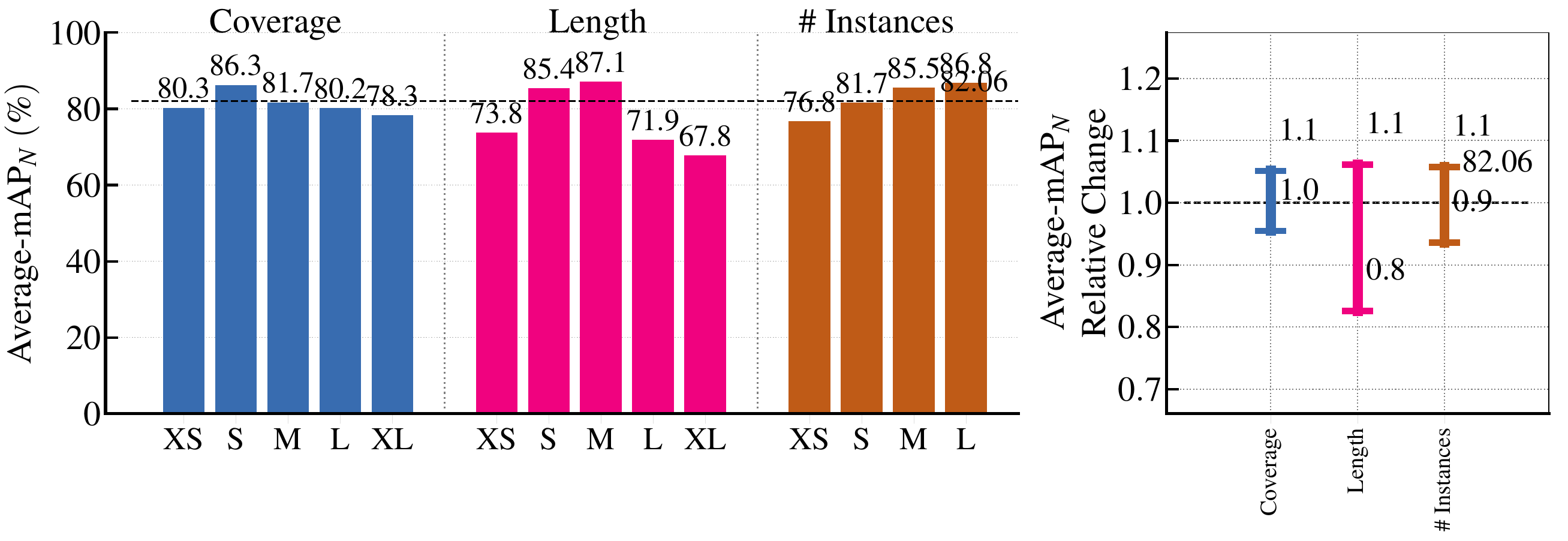} }}%
\qquad
\subfloat[\textbf{False Negative Profiling.} The false negative rates are broken down into fine-grained metrics under different coverage, length, and the number of instances.]{{\includegraphics[width=0.44\linewidth]{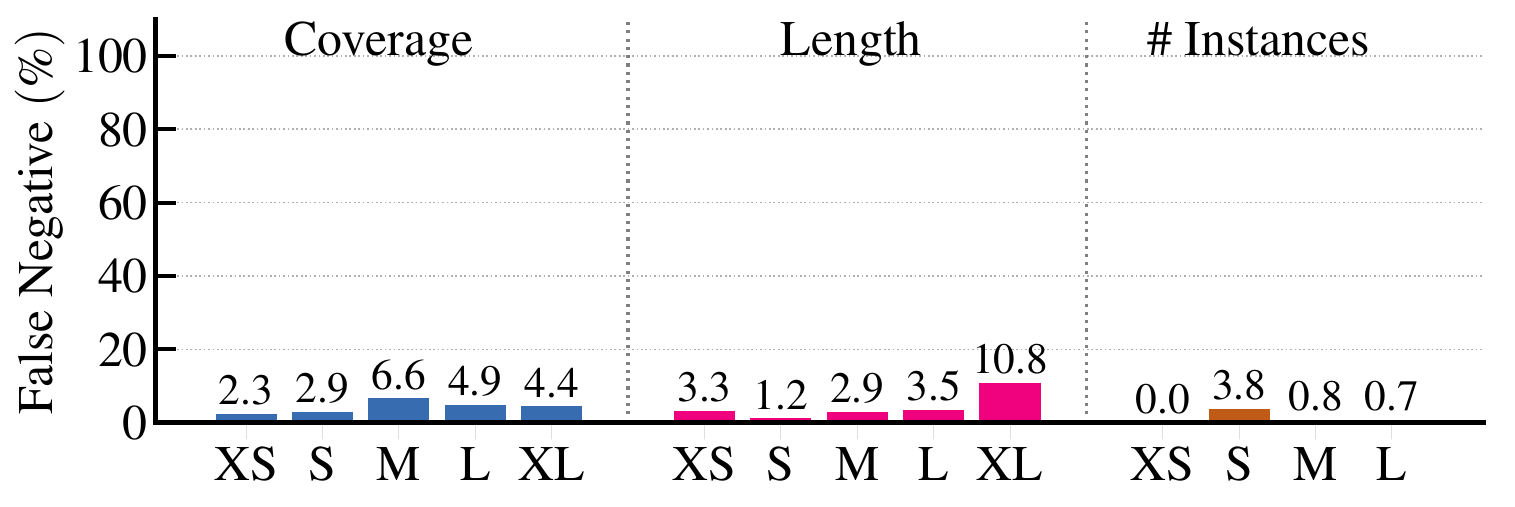} }}
\caption{We adopt the VideoMAEv2-giant as the backbone and report more error analysis of our method on THUMOS14 using~\cite{detad}. \textbf{\textit{The first row}} denotes that we use snippet features to achieve an average mAP of 69.6\%. \textbf{\textit{The second row}} denotes that we use end-to-end training by \methodname{}$^\dagger$ and achieve an average mAP of 75.4\%.}
\label{fig:error_more}
\end{figure*}

\section{Error Analysis}
\label{error}
Apart from the false positive profiling provided in the main paper, we also present the sensitive analysis and false negative (FP) profiling in Fig.~\ref{fig:error_more}. Note that the first row utilizes offline snippet features, and the second row utilizes the end-to-end training by \methodname{}$^\dagger$. For the error analysis process and metrics, we refer the readers to ~\cite{detad} for more details.

From Fig.~\ref{fig:error_more}(a), we can find that the performance across all metrics is improved by end-to-end training. Especially, the mAP of long actions (XL) is visibly increased. This phenomenon is more evident in false negative profiling. In Fig.~\ref{fig:error_more}(b), end-to-end training significantly reduces the FP rate from 24.3\% to 10.8\%, leading to better detection accuracy. Moreover, even for small actions, our method also alleviates false negative detection. For instance, it amazingly reduces the FP rate from 6.7\% to 0\% under the XS \#instances.

\section{Visualization}
\label{visual}

\begin{figure*}
\centering
{\includegraphics[trim={0cm 0 1.8cm 0},clip,width=0.47\linewidth]{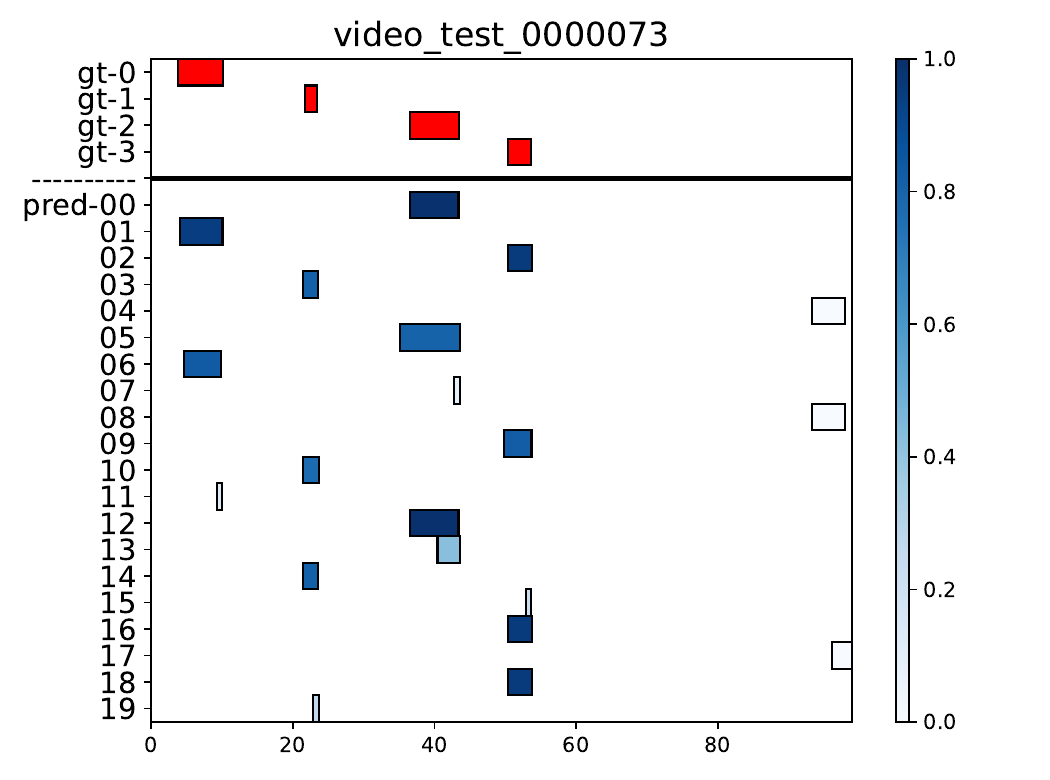} }
{\includegraphics[trim={0cm 0 1.8cm 0},clip,width=0.47\linewidth]{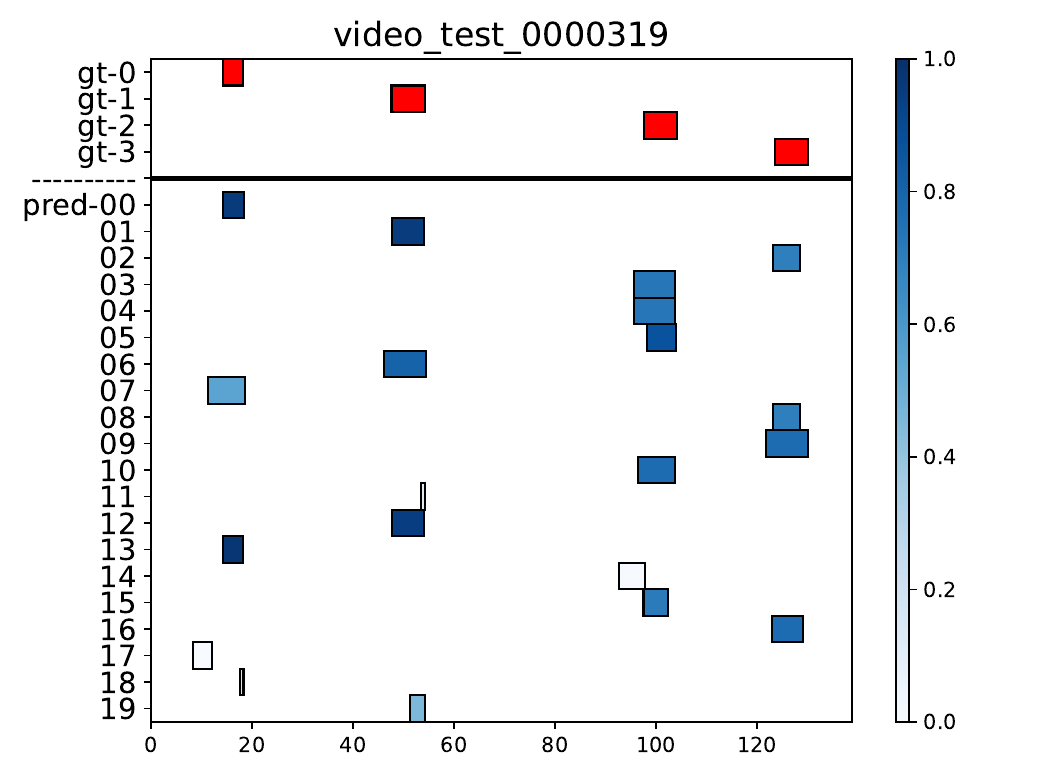} }
{\includegraphics[trim={0cm 0 1.8cm 0},clip,width=0.47\linewidth]{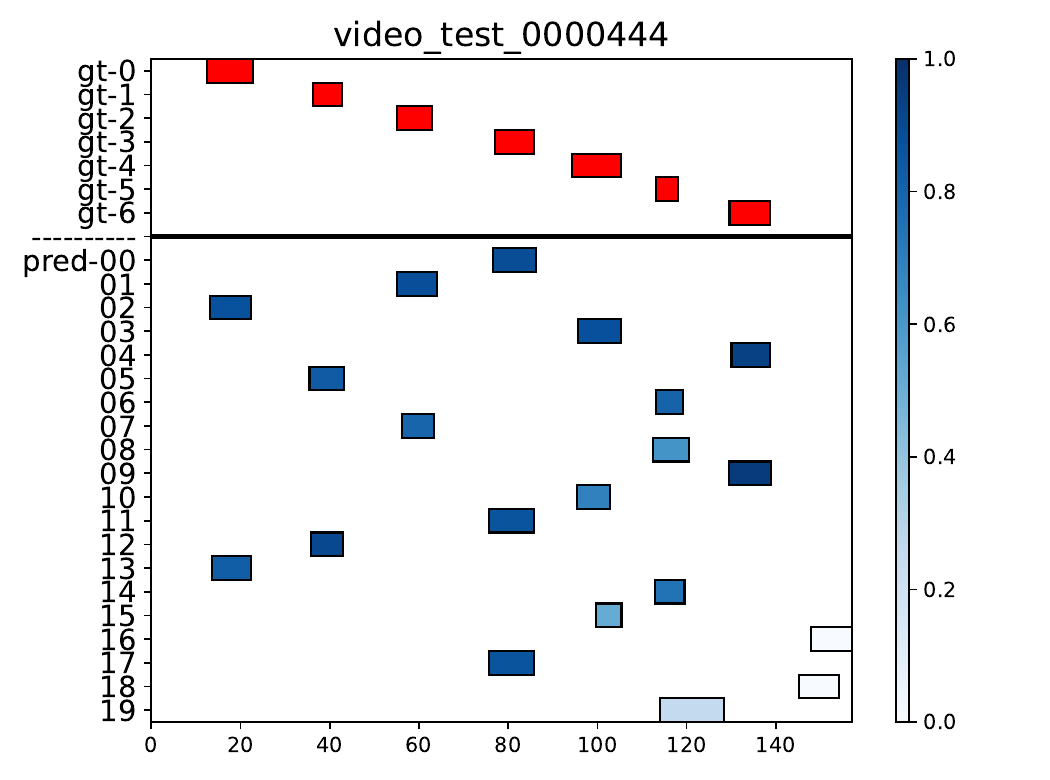} }
{\includegraphics[trim={0cm 0 1.8cm 0},clip,width=0.47\linewidth]{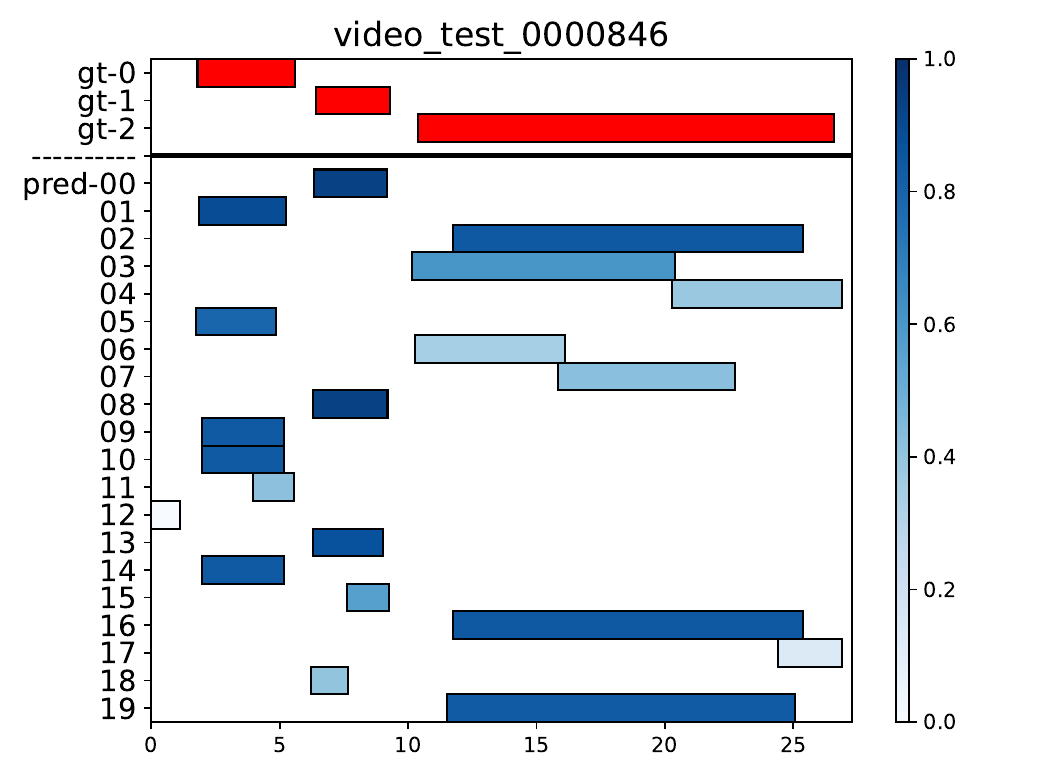} }
{\includegraphics[trim={0cm 0 1.8cm 0},clip,width=0.47\linewidth]{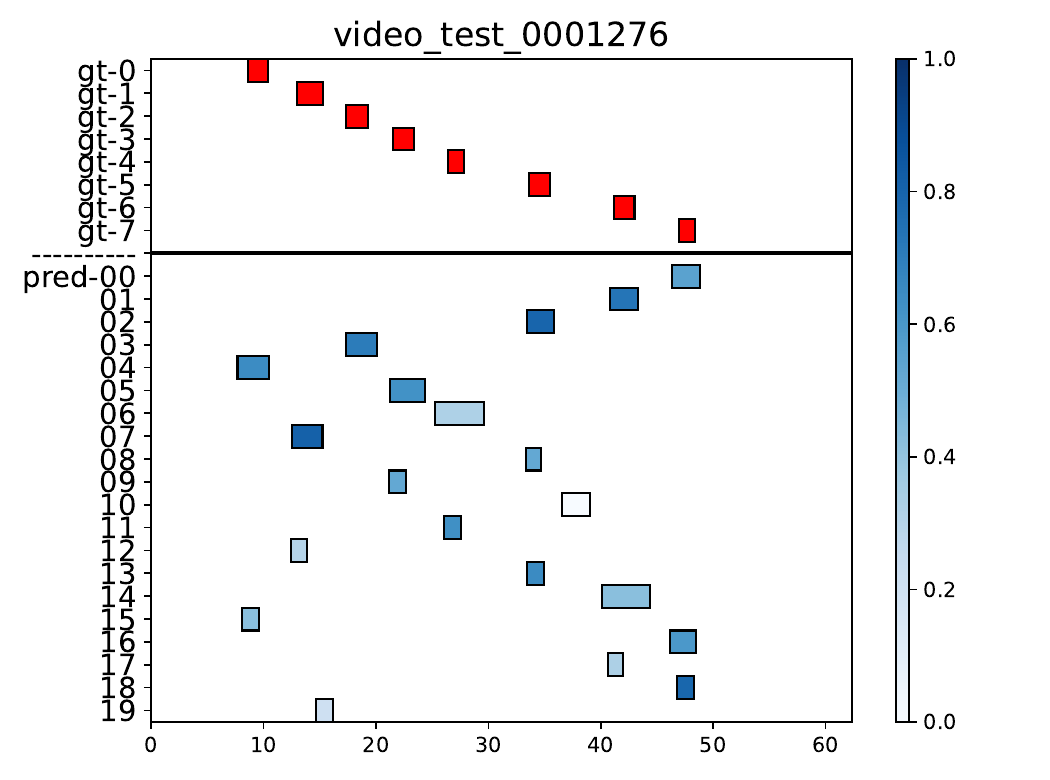} }
{\includegraphics[trim={0cm 0 1.8cm 0},clip,width=0.47\linewidth]{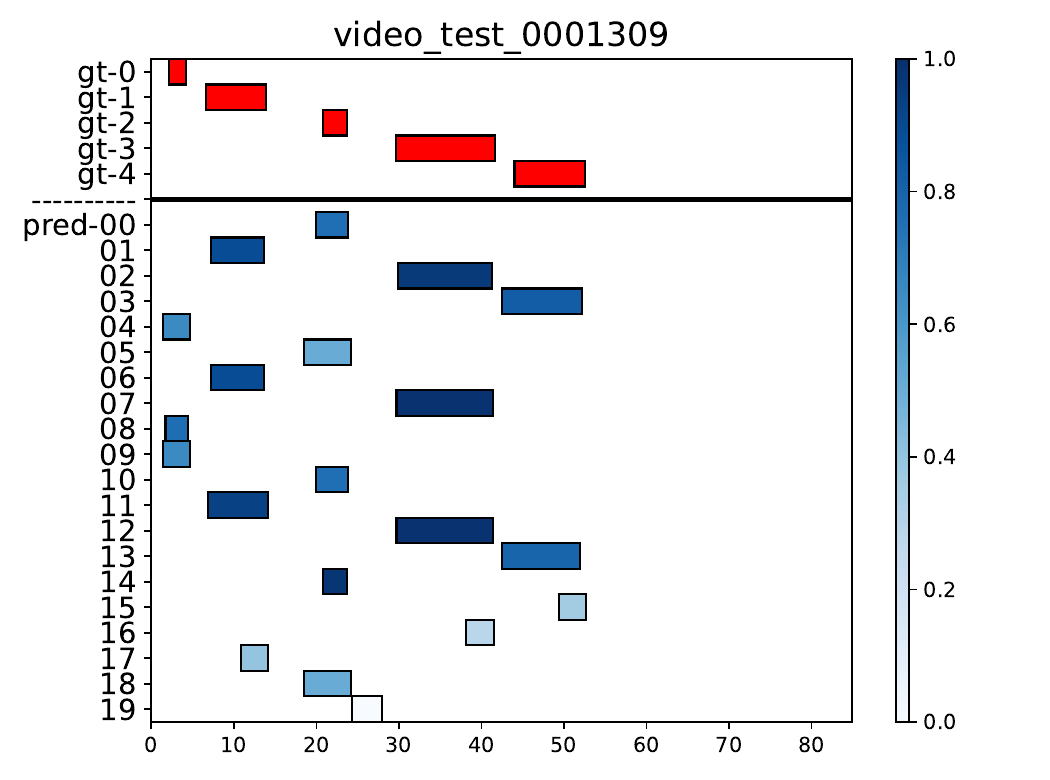} }
\caption{\textbf{Qualitative results of our method with VideoMAEv2-giant on THUMOS14.} The color of the proposal represents the maximum IoU of this proposal to ground truth actions. We plot the ground truth actions of each video (drawn in red and above the black line), and top-20 predicted proposals (drawn in colors and under the black line).}
\label{fig:qualitive_visual}
\end{figure*}

Further, we present the qualitative visualization of our prediction on THUMOS14 dataset. 
In Fig.~\ref{fig:qualitive_visual}, we plot the ground truth actions of each video (drawn in red and above the black line), and also the top-20 predicted proposals (drawn in colors and under the black line). The color of the proposal represents the maximum IoU of this proposal to the ground truth actions. Therefore, a proposal with a deeper color means it overlaps more with the ground truth, indicating this is a high-quality proposal. From the figure, we can observe that our method can yield accurate candidate actions and also provide reasonable proposal ranking.

\section{Limitations and Future Work}
\label{limitations}

One limitation of our method is how to further scale up the input data, since some datasets require extremely long video input. For instance, on Ego4D-MQ dataset, we utilize the VideoMAE-Large with 7,200 frames, costing 60GB per video. Although this is already an amazing data size for end-to-end training, however, it's meaningful to further scale up the frame resolution or model size to achieve better performance with reduced memory usage. On the other hand, loading, decoding, and processing such long videos take much longer time than pre-extracted features, which may cause difficulty in network training.

Interesting future directions include end-to-end training with multi-modality tasks, \eg, end-to-end video grounding and end-to-end moment retrieval, pretraining for action localization, and open vocabulary end-to-end temporal action detection.

{
    \small
    \bibliographystyle{ieeenat_fullname}
    \bibliography{main}

\begin{thebibliography}{79}
\providecommand{\natexlab}[1]{#1}
\providecommand{\url}[1]{\texttt{#1}}
\expandafter\ifx\csname urlstyle\endcsname\relax
  \providecommand{\doi}[1]{doi: #1}\else
  \providecommand{\doi}{doi: \begingroup \urlstyle{rm}\Url}\fi

\bibitem[Alabdulmohsin et~al.(2022)Alabdulmohsin, Neyshabur, and Zhai]{alabdulmohsin2022revisiting}
Ibrahim~M Alabdulmohsin, Behnam Neyshabur, and Xiaohua Zhai.
\newblock Revisiting neural scaling laws in language and vision.
\newblock In \emph{NeurIPS}, 2022.

\bibitem[Alwassel et~al.(2018{\natexlab{a}})Alwassel, Caba~Heilbron, Escorcia, and Ghanem]{detad}
Humam Alwassel, Fabian Caba~Heilbron, Victor Escorcia, and Bernard Ghanem.
\newblock Diagnosing error in temporal action detectors.
\newblock In \emph{ECCV}, 2018{\natexlab{a}}.

\bibitem[Alwassel et~al.(2018{\natexlab{b}})Alwassel, Heilbron, and Ghanem]{Alwassel2017ActionSS}
Humam Alwassel, Fabian~Caba Heilbron, and Bernard Ghanem.
\newblock Action search: Spotting actions in videos and its application to temporal action localization.
\newblock In \emph{ECCV}, 2018{\natexlab{b}}.

\bibitem[Bai et~al.(2020)Bai, Wang, Tong, Yang, Liu, and Liu]{Bai2020bcgnn}
Yueran Bai, Yingying Wang, Yunhai Tong, Yang Yang, Qiyue Liu, and Junhui Liu.
\newblock Boundary content graph neural network for temporal action proposal generation.
\newblock In \emph{ECCV}, 2020.

\bibitem[Brown et~al.(2020)Brown, Mann, Ryder, Subbiah, Kaplan, Dhariwal, Neelakantan, Shyam, Sastry, Askell, et~al.]{brown2020language}
Tom Brown, Benjamin Mann, Nick Ryder, Melanie Subbiah, Jared~D Kaplan, Prafulla Dhariwal, Arvind Neelakantan, Pranav Shyam, Girish Sastry, Amanda Askell, et~al.
\newblock Language models are few-shot learners.
\newblock In \emph{NeurIPS}, 2020.

\bibitem[Carreira et~al.(2019)Carreira, Noland, Hillier, and Zisserman]{carreira2019short}
Joao Carreira, Eric Noland, Chloe Hillier, and Andrew Zisserman.
\newblock A short note on the kinetics-700 human action dataset.
\newblock \emph{arXiv preprint arXiv:1907.06987}, 2019.

\bibitem[Chen et~al.(2022{\natexlab{a}})Chen, Xing, Chen, Wang, Li, Li, Liu, Wang, Zheng, Huang, et~al.]{chen2022internvideo}
Guo Chen, Sen Xing, Zhe Chen, Yi Wang, Kunchang Li, Yizhuo Li, Yi Liu, Jiahao Wang, Yin-Dong Zheng, Bingkun Huang, et~al.
\newblock Internvideo-ego4d: A pack of champion solutions to ego4d challenges.
\newblock \emph{arXiv preprint arXiv:2211.09529}, 2022{\natexlab{a}}.

\bibitem[Chen et~al.(2022{\natexlab{b}})Chen, Tao, Zhang, Wang, Ye, Wang, Hu, and Savvides]{chen2022conv}
Hao Chen, Ran Tao, Han Zhang, Yidong Wang, Wei Ye, Jindong Wang, Guosheng Hu, and Marios Savvides.
\newblock Conv-adapter: Exploring parameter efficient transfer learning for convnets.
\newblock \emph{arXiv preprint arXiv:2208.07463}, 2022{\natexlab{b}}.

\bibitem[Chen et~al.(2022{\natexlab{c}})Chen, Ge, Tong, Wang, Song, Wang, and Luo]{chen2022adaptformer}
Shoufa Chen, Chongjian Ge, Zhan Tong, Jiangliu Wang, Yibing Song, Jue Wang, and Ping Luo.
\newblock Adaptformer: Adapting vision transformers for scalable visual recognition.
\newblock In \emph{NeurIPS}, 2022{\natexlab{c}}.

\bibitem[Chen et~al.(2016)Chen, Xu, Zhang, and Guestrin]{chen2016training}
Tianqi Chen, Bing Xu, Chiyuan Zhang, and Carlos Guestrin.
\newblock Training deep nets with sublinear memory cost.
\newblock \emph{arXiv preprint arXiv:1604.06174}, 2016.

\bibitem[Chen et~al.(2023)Chen, Qian, Tang, Lai, Liu, Han, and Jia]{chen2023longlora}
Yukang Chen, Shengju Qian, Haotian Tang, Xin Lai, Zhijian Liu, Song Han, and Jiaya Jia.
\newblock Longlora: Efficient fine-tuning of long-context large language models.
\newblock \emph{arXiv preprint arXiv:2309.12307}, 2023.

\bibitem[Cheng and Bertasius(2022)]{cheng2022tallformer}
Feng Cheng and Gedas Bertasius.
\newblock Tallformer: Temporal action localization with long-memory transformer.
\newblock In \emph{ECCV}, 2022.

\bibitem[Cheng et~al.(2023)Cheng, Xu, Xiong, Chen, Li, Li, and Xia]{cheng2022stochastic}
Feng Cheng, Mingze Xu, Yuanjun Xiong, Hao Chen, Xinyu Li, Wei Li, and Wei Xia.
\newblock Stochastic backpropagation: A memory efficient strategy for training video models.
\newblock In \emph{CVPR}, 2023.

\bibitem[Contributors(2020)]{2020mmaction2}
MMAction2 Contributors.
\newblock Openmmlab's next generation video understanding toolbox and benchmark.
\newblock \url{https://github.com/open-mmlab/mmaction2}, 2020.

\bibitem[Damen et~al.(2018)Damen, Doughty, Farinella, Fidler, Furnari, Kazakos, Moltisanti, Munro, Perrett, Price, et~al.]{damen2018scaling}
Dima Damen, Hazel Doughty, Giovanni~Maria Farinella, Sanja Fidler, Antonino Furnari, Evangelos Kazakos, Davide Moltisanti, Jonathan Munro, Toby Perrett, Will Price, et~al.
\newblock Scaling egocentric vision: The epic-kitchens dataset.
\newblock In \emph{ECCV}, 2018.

\bibitem[Dao et~al.(2022)Dao, Fu, Ermon, Rudra, and R{\'e}]{dao2022flashattention}
Tri Dao, Dan Fu, Stefano Ermon, Atri Rudra, and Christopher R{\'e}.
\newblock Flashattention: Fast and memory-efficient exact attention with io-awareness.
\newblock In \emph{NeurIPS}, 2022.

\bibitem[Dosovitskiy et~al.(2021)Dosovitskiy, Beyer, Kolesnikov, Weissenborn, Zhai, Unterthiner, Dehghani, Minderer, Heigold, Gelly, et~al.]{dosovitskiy2020image}
Alexey Dosovitskiy, Lucas Beyer, Alexander Kolesnikov, Dirk Weissenborn, Xiaohua Zhai, Thomas Unterthiner, Mostafa Dehghani, Matthias Minderer, Georg Heigold, Sylvain Gelly, et~al.
\newblock An image is worth 16x16 words: Transformers for image recognition at scale.
\newblock In \emph{ICLR}, 2021.

\bibitem[Escorcia et~al.(2016)Escorcia, Heilbron, Niebles, and Ghanem]{escorcia2016DAPsDA}
Victor Escorcia, Fabian~Caba Heilbron, Juan~Carlos Niebles, and Bernard Ghanem.
\newblock {DAPs}: Deep action proposals for action understanding.
\newblock In \emph{ECCV}, 2016.

\bibitem[Fan et~al.(2021)Fan, Xiong, Mangalam, Li, Yan, Malik, and Feichtenhofer]{fan2021multiscale}
Haoqi Fan, Bo Xiong, Karttikeya Mangalam, Yanghao Li, Zhicheng Yan, Jitendra Malik, and Christoph Feichtenhofer.
\newblock Multiscale vision transformers.
\newblock In \emph{ICCV}, 2021.

\bibitem[Feichtenhofer et~al.(2019)Feichtenhofer, Fan, Malik, and He]{slowfast_iccv19}
Christoph Feichtenhofer, Haoqi Fan, Jitendra Malik, and Kaiming He.
\newblock {SlowFast} networks for video recognition.
\newblock In \emph{ICCV}, 2019.

\bibitem[Grauman et~al.(2022)Grauman, Westbury, Byrne, Chavis, Furnari, Girdhar, Hamburger, Jiang, Liu, and Liu]{grauman2022ego4d}
Kristen Grauman, Andrew Westbury, Eugene Byrne, Zachary Chavis, Antonino Furnari, Rohit Girdhar, Jackson Hamburger, Hao Jiang, Miao Liu, and et~al. Liu, Xingyu.
\newblock Ego4d: Around the world in 3,000 hours of egocentric video.
\newblock In \emph{CVPR}, 2022.

\bibitem[He et~al.(2022)He, Chen, Xie, Li, Doll{\'a}r, and Girshick]{he2022masked}
Kaiming He, Xinlei Chen, Saining Xie, Yanghao Li, Piotr Doll{\'a}r, and Ross Girshick.
\newblock Masked autoencoders are scalable vision learners.
\newblock In \emph{CVPR}, 2022.

\bibitem[Heilbron et~al.(2015)Heilbron, Escorcia, Ghanem, and Niebles]{caba2015activitynet}
Fabian~Caba Heilbron, Victor Escorcia, Bernard Ghanem, and Juan~Carlos Niebles.
\newblock {ActivityNet}: A large-scale video benchmark for human activity understanding.
\newblock In \emph{CVPR}, 2015.

\bibitem[Hendrycks and Gimpel(2016)]{hendrycks2016gaussian}
Dan Hendrycks and Kevin Gimpel.
\newblock Gaussian error linear units (gelus).
\newblock \emph{arXiv preprint arXiv:1606.08415}, 2016.

\bibitem[Houlsby et~al.(2019)Houlsby, Giurgiu, Jastrzebski, Morrone, De~Laroussilhe, Gesmundo, Attariyan, and Gelly]{houlsby2019parameter}
Neil Houlsby, Andrei Giurgiu, Stanislaw Jastrzebski, Bruna Morrone, Quentin De~Laroussilhe, Andrea Gesmundo, Mona Attariyan, and Sylvain Gelly.
\newblock Parameter-efficient transfer learning for nlp.
\newblock In \emph{ICML}, 2019.

\bibitem[Hu et~al.(2021)Hu, Shen, Wallis, Allen-Zhu, Li, Wang, Wang, and Chen]{hu2021lora}
Edward~J Hu, Yelong Shen, Phillip Wallis, Zeyuan Allen-Zhu, Yuanzhi Li, Shean Wang, Lu Wang, and Weizhu Chen.
\newblock Lora: Low-rank adaptation of large language models.
\newblock In \emph{ICML}, 2021.

\bibitem[Jia et~al.(2022)Jia, Tang, Chen, Cardie, Belongie, Hariharan, and Lim]{jia2022visual}
Menglin Jia, Luming Tang, Bor-Chun Chen, Claire Cardie, Serge Belongie, Bharath Hariharan, and Ser-Nam Lim.
\newblock Visual prompt tuning.
\newblock In \emph{ECCV}, 2022.

\bibitem[Jiang et~al.(2014)Jiang, Liu, Zamir, Toderici, Laptev, Shah, and Sukthankar]{jiang2014thumos}
YG Jiang, J Liu, A~Roshan Zamir, G Toderici, I Laptev, M Shah, and R Sukthankar.
\newblock Thumos challenge: Action recognition with a large number of classes, 2014.

\bibitem[Kay et~al.(2017)Kay, Carreira, Simonyan, Zhang, Hillier, Vijayanarasimhan, Viola, Green, Back, Natsev, et~al.]{kay2017kinetics}
Will Kay, Joao Carreira, Karen Simonyan, Brian Zhang, Chloe Hillier, Sudheendra Vijayanarasimhan, Fabio Viola, Tim Green, Trevor Back, Paul Natsev, et~al.
\newblock The kinetics human action video dataset.
\newblock \emph{arXiv preprint arXiv:1705.06950}, 2017.

\bibitem[Lester et~al.(2021)Lester, Al-Rfou, and Constant]{lester2021power}
Brian Lester, Rami Al-Rfou, and Noah Constant.
\newblock The power of scale for parameter-efficient prompt tuning.
\newblock In \emph{EMNLP}, 2021.

\bibitem[Li and Liang(2021)]{li2021prefix}
Xiang~Lisa Li and Percy Liang.
\newblock Prefix-tuning: Optimizing continuous prompts for generation.
\newblock In \emph{ACL}, 2021.

\bibitem[Lin et~al.(2021)Lin, Xu, Luo, Wang, Tai, Wang, Li, Huang, and Fu]{lin2021learning}
Chuming Lin, Chengming Xu, Donghao Luo, Yabiao Wang, Ying Tai, Chengjie Wang, Jilin Li, Feiyue Huang, and Yanwei Fu.
\newblock Learning salient boundary feature for anchor-free temporal action localization.
\newblock In \emph{CVPR}, 2021.

\bibitem[Lin et~al.(2019)Lin, Liu, Li, Ding, and Wen]{lin2019bmn}
Tianwei Lin, Xiao Liu, Xin Li, Errui Ding, and Shilei Wen.
\newblock {BMN:} boundary-matching network for temporal action proposal generation.
\newblock In \emph{ICCV}, 2019.

\bibitem[Liu and Wang(2020)]{liu2020progressive}
Qinying Liu and Zilei Wang.
\newblock Progressive boundary refinement network for temporal action detection.
\newblock In \emph{AAAI}, 2020.

\bibitem[Liu et~al.(2023)Liu, Xu, Zhao, Zhao, and Ghanem]{liu2023etad}
Shuming Liu, Mengmeng Xu, Chen Zhao, Xu Zhao, and Bernard Ghanem.
\newblock Etad: Training action detection end to end on a laptop.
\newblock In \emph{CVPRW}, 2023.

\bibitem[Liu et~al.(2022{\natexlab{a}})Liu, Bai, and Bai]{liu2022empirical}
Xiaolong Liu, Song Bai, and Xiang Bai.
\newblock An empirical study of end-to-end temporal action detection.
\newblock In \emph{CVPR}, 2022{\natexlab{a}}.

\bibitem[Liu et~al.(2022{\natexlab{b}})Liu, Wang, Hu, Tang, Zhang, Bai, and Bai]{liu2022end}
Xiaolong Liu, Qimeng Wang, Yao Hu, Xu Tang, Shiwei Zhang, Song Bai, and Xiang Bai.
\newblock End-to-end temporal action detection with transformer.
\newblock \emph{IEEE Transactions on Image Processing}, 31:\penalty0 5427--5441, 2022{\natexlab{b}}.

\bibitem[Liu et~al.(2021)Liu, Lin, Cao, Hu, Wei, Zhang, Lin, and Guo]{liu2021swin}
Ze Liu, Yutong Lin, Yue Cao, Han Hu, Yixuan Wei, Zheng Zhang, Stephen Lin, and Baining Guo.
\newblock Swin transformer: Hierarchical vision transformer using shifted windows.
\newblock In \emph{ICCV}, 2021.

\bibitem[Liu et~al.(2022{\natexlab{c}})Liu, Ning, Cao, Wei, Zhang, Lin, and Hu]{liu2022video}
Ze Liu, Jia Ning, Yue Cao, Yixuan Wei, Zheng Zhang, Stephen Lin, and Han Hu.
\newblock Video swin transformer.
\newblock In \emph{Proceedings of the IEEE/CVF conference on computer vision and pattern recognition}, pages 3202--3211, 2022{\natexlab{c}}.

\bibitem[Mai et~al.(2023)Mai, Hamdi, Giancola, Zhao, and Ghanem]{mai2023egoloc}
Jinjie Mai, Abdullah Hamdi, Silvio Giancola, Chen Zhao, and Bernard Ghanem.
\newblock Egoloc: Revisiting 3d object localization from egocentric videos with visual queries.
\newblock In \emph{Proceedings of the IEEE/CVF International Conference on Computer Vision}, pages 45--57, 2023.

\bibitem[Micikevicius et~al.(2017)Micikevicius, Narang, Alben, Diamos, Elsen, Garcia, Ginsburg, Houston, Kuchaiev, Venkatesh, et~al.]{micikevicius2017mixed}
Paulius Micikevicius, Sharan Narang, Jonah Alben, Gregory Diamos, Erich Elsen, David Garcia, Boris Ginsburg, Michael Houston, Oleksii Kuchaiev, Ganesh Venkatesh, et~al.
\newblock Mixed precision training.
\newblock In \emph{ICLR}, 2017.

\bibitem[Mun et~al.(2020)Mun, Cho, and Han]{Mun_2020_CVPR}
Jonghwan Mun, Minsu Cho, and Bohyung Han.
\newblock Local-global video-text interactions for temporal grounding.
\newblock In \emph{CVPR}, 2020.

\bibitem[OpenAI(2023)]{openai2023gpt4}
OpenAI.
\newblock Gpt-4 technical report, 2023.

\bibitem[Qing et~al.(2021)Qing, Su, Gan, Wang, Wu, Wang, Qiao, Yan, Gao, and Sang]{qing2021temporal}
Zhiwu Qing, Haisheng Su, Weihao Gan, Dongliang Wang, Wei Wu, Xiang Wang, Yu Qiao, Junjie Yan, Changxin Gao, and Nong Sang.
\newblock Temporal context aggregation network for temporal action proposal refinement.
\newblock In \emph{CVPR}, 2021.

\bibitem[Radford et~al.(2018)Radford, Narasimhan, Salimans, Sutskever, et~al.]{radford2018improving}
Alec Radford, Karthik Narasimhan, Tim Salimans, Ilya Sutskever, et~al.
\newblock Improving language understanding by generative pre-training.
\newblock \emph{OpenAI}, 2018.

\bibitem[Radford et~al.(2019)Radford, Wu, Child, Luan, Amodei, Sutskever, et~al.]{radford2019language}
Alec Radford, Jeffrey Wu, Rewon Child, David Luan, Dario Amodei, Ilya Sutskever, et~al.
\newblock Language models are unsupervised multitask learners.
\newblock \emph{OpenAI blog}, 1\penalty0 (8):\penalty0 9, 2019.

\bibitem[Radford et~al.(2021)Radford, Kim, Hallacy, Ramesh, Goh, Agarwal, Sastry, Askell, Mishkin, Clark, et~al.]{radford2021learning}
Alec Radford, Jong~Wook Kim, Chris Hallacy, Aditya Ramesh, Gabriel Goh, Sandhini Agarwal, Girish Sastry, Amanda Askell, Pamela Mishkin, Jack Clark, et~al.
\newblock Learning transferable visual models from natural language supervision.
\newblock In \emph{ICML}, 2021.

\bibitem[Ramazanova et~al.(2023)Ramazanova, Escorcia, Heilbron, Zhao, and Ghanem]{ramazanova2023owl}
Merey Ramazanova, Victor Escorcia, Fabian~Caba Heilbron, Chen Zhao, and Bernard Ghanem.
\newblock Owl (observe, watch, listen): Localizing actions in egocentric video via audiovisual temporal context.
\newblock \emph{Proceedings of the IEEE/CVF Conference on Computer Vision and Pattern Recognition Workshop (CVPRW)}, 2023.

\bibitem[Shao et~al.(2023)Shao, Wang, Quan, Zheng, Yang, and Yang]{shao2023action}
Jiayi Shao, Xiaohan Wang, Ruijie Quan, Junjun Zheng, Jiang Yang, and Yi Yang.
\newblock Action sensitivity learning for temporal action localization.
\newblock In \emph{ICCV}, 2023.

\bibitem[Shi et~al.(2022)Shi, Zhong, Cao, Zhang, Ma, Li, and Tao]{shi2022react}
Dingfeng Shi, Yujie Zhong, Qiong Cao, Jing Zhang, Lin Ma, Jia Li, and Dacheng Tao.
\newblock React: Temporal action detection with relational queries.
\newblock In \emph{ECCV}, 2022.

\bibitem[Shi et~al.(2023)Shi, Zhong, Cao, Ma, Li, and Tao]{shi2023tridet}
Dingfeng Shi, Yujie Zhong, Qiong Cao, Lin Ma, Jia Li, and Dacheng Tao.
\newblock Tridet: Temporal action detection with relative boundary modeling.
\newblock In \emph{CVPR}, 2023.

\bibitem[Soldan et~al.(2021)Soldan, Xu, Qu, Tegner, and Ghanem]{Soldan_2021_ICCV}
Mattia Soldan, Mengmeng Xu, Sisi Qu, Jesper Tegner, and Bernard Ghanem.
\newblock {VLG-Net}: Video-language graph matching network for video grounding.
\newblock In \emph{ICCVW}, 2021.

\bibitem[Soldan et~al.(2022)Soldan, Pardo, Alc{\'a}zar, Heilbron, Zhao, Giancola, and Ghanem]{soldan2021mad}
Mattia Soldan, Alejandro Pardo, Juan~Le{\'o}n Alc{\'a}zar, Fabian~Caba Heilbron, Chen Zhao, Silvio Giancola, and Bernard Ghanem.
\newblock Mad: A scalable dataset for language grounding in videos from movie audio descriptions.
\newblock \emph{Proceedings of the IEEE/CVF Conference on Computer Vision and Pattern Recognition (CVPR)}, 2022.

\bibitem[Sui et~al.(2023)Sui, Mu, and Li]{sui2023nms}
Lin Sui, Fangzhou Mu, and Yin Li.
\newblock Nms threshold matters for ego4d moment queries--2nd place solution to the ego4d moment queries challenge 2023.
\newblock \emph{arXiv preprint arXiv:2307.02025}, 2023.

\bibitem[Sung et~al.(2022{\natexlab{a}})Sung, Cho, and Bansal]{sung2022lst}
Yi-Lin Sung, Jaemin Cho, and Mohit Bansal.
\newblock Lst: Ladder side-tuning for parameter and memory efficient transfer learning.
\newblock In \emph{NeurIPS}, 2022{\natexlab{a}}.

\bibitem[Sung et~al.(2022{\natexlab{b}})Sung, Cho, and Bansal]{sung2022vl}
Yi-Lin Sung, Jaemin Cho, and Mohit Bansal.
\newblock Vl-adapter: Parameter-efficient transfer learning for vision-and-language tasks.
\newblock In \emph{CVPR}, 2022{\natexlab{b}}.

\bibitem[Tan et~al.(2021)Tan, Tang, Wang, and Wu]{tan2021relaxed}
Jing Tan, Jiaqi Tang, Limin Wang, and Gangshan Wu.
\newblock Relaxed transformer decoders for direct action proposal generation.
\newblock In \emph{ICCV}, 2021.

\bibitem[Tang et~al.(2023)Tang, Kim, and Sohn]{tang2023temporalmaxer}
Tuan~N Tang, Kwonyoung Kim, and Kwanghoon Sohn.
\newblock Temporalmaxer: Maximize temporal context with only max pooling for temporal action localization.
\newblock \emph{arXiv preprint arXiv:2303.09055}, 2023.

\bibitem[Tong et~al.(2022)Tong, Song, Wang, and Wang]{tong2022videomae}
Zhan Tong, Yibing Song, Jue Wang, and Limin Wang.
\newblock Videomae: Masked autoencoders are data-efficient learners for self-supervised video pre-training.
\newblock In \emph{NeurIPS}, 2022.

\bibitem[Vaswani et~al.(2017)Vaswani, Shazeer, Parmar, Uszkoreit, Jones, Gomez, Kaiser, and Polosukhin]{vaswani2017attention}
Ashish Vaswani, Noam Shazeer, Niki Parmar, Jakob Uszkoreit, Llion Jones, Aidan~N Gomez, {\L}ukasz Kaiser, and Illia Polosukhin.
\newblock Attention is all you need.
\newblock In \emph{NeurIPS}, 2017.

\bibitem[Wang et~al.(2021)Wang, Cai, Zou, and Xiong]{wang2021rgb}
Chenhao Wang, Hongxiang Cai, Yuxin Zou, and Yichao Xiong.
\newblock Rgb stream is enough for temporal action detection.
\newblock \emph{arXiv preprint arXiv:2107.04362}, 2021.

\bibitem[Wang et~al.(2023)Wang, Huang, Zhao, Tong, He, Wang, Wang, and Qiao]{wang2023videomae}
Limin Wang, Bingkun Huang, Zhiyu Zhao, Zhan Tong, Yinan He, Yi Wang, Yali Wang, and Yu Qiao.
\newblock Videomae v2: Scaling video masked autoencoders with dual masking.
\newblock In \emph{ICCV}, 2023.

\bibitem[Wang et~al.(2022)Wang, Li, Li, He, Huang, Zhao, Zhang, Xu, Liu, Wang, et~al.]{wang2022internvideo}
Yi Wang, Kunchang Li, Yizhuo Li, Yinan He, Bingkun Huang, Zhiyu Zhao, Hongjie Zhang, Jilan Xu, Yi Liu, Zun Wang, et~al.
\newblock Internvideo: General video foundation models via generative and discriminative learning.
\newblock \emph{arXiv preprint arXiv:2212.03191}, 2022.

\bibitem[Wang et~al.(2020)Wang, Gao, Wang, Li, and Wu]{wang2020boundary}
Zhenzhi Wang, Ziteng Gao, Limin Wang, Zhifeng Li, and Gangshan Wu.
\newblock Boundary-aware cascade networks for temporal action segmentation.
\newblock In \emph{ECCV}, 2020.

\bibitem[Xia et~al.(2022)Xia, Wang, Zhou, Zheng, and Tang]{xia2022learning}
Kun Xia, Le Wang, Sanping Zhou, Nanning Zheng, and Wei Tang.
\newblock Learning to refactor action and co-occurrence features for temporal action localization.
\newblock In \emph{CVPR}, 2022.

\bibitem[Xu et~al.(2020)Xu, Zhao, Rojas, Thabet, and Ghanem]{xu2020g}
Mengmeng Xu, Chen Zhao, David~S Rojas, Ali Thabet, and Bernard Ghanem.
\newblock {G-TAD}: Sub-graph localization for temporal action detection.
\newblock In \emph{CVPR}, 2020.

\bibitem[Yang et~al.(2020)Yang, Peng, Zhang, Fu, and Han]{yang2020revisiting}
Le Yang, Houwen Peng, Dingwen Zhang, Jianlong Fu, and Junwei Han.
\newblock Revisiting anchor mechanisms for temporal action localization.
\newblock \emph{IEEE Transactions on Image Processing}, 29:\penalty0 8535--8548, 2020.

\bibitem[Yang et~al.(2023{\natexlab{a}})Yang, Chen, Zheng, Lu, and Wang]{yang2023basictad}
Min Yang, Guo Chen, Yin-Dong Zheng, Tong Lu, and Limin Wang.
\newblock Basictad: an astounding rgb-only baseline for temporal action detection.
\newblock \emph{Computer Vision and Image Understanding}, 232:\penalty0 103692, 2023{\natexlab{a}}.

\bibitem[Yang et~al.(2023{\natexlab{b}})Yang, Zhu, Xie, Zhang, Chen, and Li]{yang2023aim}
Taojiannan Yang, Yi Zhu, Yusheng Xie, Aston Zhang, Chen Chen, and Mu Li.
\newblock Aim: Adapting image models for efficient video action recognition.
\newblock In \emph{ICLR}, 2023{\natexlab{b}}.

\bibitem[Yao et~al.(2016)Yao, Mei, and Rui]{yao2016highlight}
Ting Yao, Tao Mei, and Yong Rui.
\newblock Highlight detection with pairwise deep ranking for first-person video summarization.
\newblock In \emph{CVPR}, 2016.

\bibitem[Yin et~al.(2023)Yin, Han, Li, Feng, and Bai]{yin2023parameter}
Dongshuo Yin, Xueting Han, Bin Li, Hao Feng, and Jing Bai.
\newblock Parameter-efficient is not sufficient: Exploring parameter, memory, and time efficient adapter tuning for dense predictions.
\newblock \emph{arXiv preprint arXiv:2306.09729}, 2023.

\bibitem[Zhang et~al.(2022)Zhang, Wu, and Li]{zhang2022actionformer}
Chenlin Zhang, Jianxin Wu, and Yin Li.
\newblock Actionformer: Localizing moments of actions with transformers.
\newblock In \emph{ECCV}, 2022.

\bibitem[Zhao et~al.(2021)Zhao, Thabet, and Ghanem]{zhao2021video}
Chen Zhao, Ali~K Thabet, and Bernard Ghanem.
\newblock Video self-stitching graph network for temporal action localization.
\newblock In \emph{ICCV}, 2021.

\bibitem[Zhao et~al.(2022)Zhao, Ramazanova, Xu, and Ghanem]{zhao2022segtad}
Chen Zhao, Merey Ramazanova, Mengmeng Xu, and Bernard Ghanem.
\newblock Segtad: Precise temporal action detection via semantic segmentation.
\newblock In \emph{ECCVW}, 2022.

\bibitem[Zhao et~al.(2023{\natexlab{a}})Zhao, Liu, Mangalam, and Ghanem]{zhao2023re2tal}
Chen Zhao, Shuming Liu, Karttikeya Mangalam, and Bernard Ghanem.
\newblock Re$^2${TAL}: Rewiring pretrained video backbones for reversible temporal action localization.
\newblock In \emph{CVPR}, 2023{\natexlab{a}}.

\bibitem[Zhao et~al.(2024)Zhao, Liu, Mangalam, Qian, Zohra, Alghannam, Malik, and Ghanem]{zhao2024dr2net}
Chen Zhao, Shuming Liu, Karttikeya Mangalam, Guocheng Qian, Fatimah Zohra, Abdulmohsen Alghannam, Jitendra Malik, and Bernard Ghanem.
\newblock Dr$^2${Net}: Dynamic reversible dual-residual networks for memory-efficient finetuning.
\newblock In \emph{Proceedings of the IEEE/CVF Conference on Computer Vision and Pattern Recognition (CVPR)}, 2024.

\bibitem[Zhao et~al.(2019)Zhao, Yan, Torresani, and Torralba]{zhao2019hacs}
Hang Zhao, Zhicheng Yan, Lorenzo Torresani, and Antonio Torralba.
\newblock {HACS}: Human action clips and segments dataset for recognition and temporal localization.
\newblock \emph{ICCV}, 2019.

\bibitem[Zhao et~al.(2017)Zhao, Zhang, Wu, Yang, Zhou, Yan, Wang, Xiong, Lin, Qiao, et~al.]{zhao2017cuhk}
Y Zhao, B Zhang, Z Wu, S Yang, L Zhou, S Yan, L Wang, Y Xiong, D Lin, Y Qiao, et~al.
\newblock Cuhk \& ethz \& siat submission to activitynet challenge 2017.
\newblock \emph{CVPR ActivityNet Workshop}, 2017.

\bibitem[Zhao et~al.(2023{\natexlab{b}})Zhao, Wang, and Zhao]{Zhao_2023_ICCV}
Zixuan Zhao, Dongqi Wang, and Xu Zhao.
\newblock Movement enhancement toward multi-scale video feature representation for temporal action detection.
\newblock In \emph{Proceedings of the IEEE/CVF International Conference on Computer Vision (ICCV)}, pages 13555--13564, 2023{\natexlab{b}}.

\end{thebibliography}
}

\end{document}